\pdfoutput=1
\documentclass{article}
\usepackage[utf8]{inputenc}
\usepackage{geometry}
\usepackage{graphicx}
\usepackage{bm}
\usepackage{amsmath}
\usepackage[normalem]{ulem}
\usepackage{graphicx}
\usepackage{color,soul}
\usepackage[colorlinks=true, citecolor=black, urlcolor=blue]{hyperref}
\usepackage{natbib}
\usepackage{tabularx}
\usepackage{graphicx}
\usepackage{adjustbox}
\usepackage{geometry}
\usepackage{authblk}
\usepackage{tabularx}
\usepackage[table,xcdraw]{xcolor}
\usepackage{textcomp}
\usepackage{siunitx}
\usepackage{subfig}
\usepackage{float}
\usepackage{multirow} 
\usepackage{hyperref}
\usepackage{amssymb}
\usepackage{makecell}

\usepackage{geometry}
\usepackage{authblk}
\usepackage[colorlinks=true, citecolor=black, urlcolor=blue]{hyperref}

\begin{document}

\title{Physics-Informed Tree Search for High-Dimensional Computational Design} 



\author[1,2,*]{Suvo Banik}
\author[1,2]{Troy D. Loeffler}
\author[1]{Henry Chan}
\author[1,2]{Sukriti Manna}
\author[3]{Orcun Yildiz}
\author[3]{Tom Peterka}
\author[1,2,*]{Subramanian Sankaranarayanan}

\affil[1]{Department of Mechanical and Industrial Engineering, University of Illinois, Chicago, Illinois 60607, United States}
\affil[2]{Center for Nanoscale Materials, Argonne National Laboratory, Lemont, Illinois 60439, United States}
\affil[3]{Mathematics and Computer Science Division, Argonne National Laboratory, Lemont, Illinois 60439, United States}
\affil[*]{Corresponding authors: \href{mailto:skrssank@uic.edu}{skrssank@uic.edu}, \href{mailto:sbanik2@anl.gov}{sbanik2@anl.gov};}

\maketitle
\begin{abstract}
 
High-dimensional design spaces underpin a wide range of physics-based modeling and computational design tasks in science and engineering. These problems are commonly formulated as constrained black-box searches over rugged objective landscapes, where function evaluations are expensive and gradients are unavailable or unreliable. Conventional global search engines and optimizers struggle in such settings due to the exponential scaling of design spaces, the presence of multiple local basins, and the absence of physical guidance in sampling. We present a physics-informed Monte Carlo Tree Search (MCTS) framework that extends policy-driven tree-based reinforcement concepts to continuous, high-dimensional scientific optimization. Our method integrates population-level decision trees with surrogate-guided directional sampling, reward shaping, and hierarchical switching between global exploration and local exploitation. These ingredients allows efficient traversal of non-convex, multimodal landscapes where physically meaningful optima are sparse. We benchmark our approach against standard global optimization baselines on a suite of canonical test functions, demonstrating superior or comparable performance in terms of convergence, robustness, and generalization. Beyond synthetic tests, we demonstrate physics-consistent applicability to (i) crystal structure optimization from clusters to bulk, (ii) fitting of classical interatomic potentials from quantum datasets, and (iii) constrained engineering design problems governed by continuum mechanics. Across all cases, the method converges with high fidelity and evaluation efficiency while preserving physical constraints. Overall, our work establishes physics-informed tree search as a scalable and interpretable paradigm for computational design and high-dimensional scientific optimization, bridging discrete decision-making frameworks with continuous black-box search in scientific design workflows.

\end{abstract}

\newpage

\section{Introduction}

Computational design frequently involves high-dimensional optimization in continuous search spaces, where the objective encodes a physical cost, performance metric, or stability criterion evaluated through a simulator or numerical model\cite{allaire2007numerical}. Such problems arise in materials design\cite{zunger2018inverse}, inverse structure search\cite{oganov2011evolutionary}, potential fitting\cite{martinez2013fitting}, topology and mechanical design\cite{zhu2016topology}, and multi-physics co-optimization\cite{hennigh2021nvidia}—settings where the objective function is effectively a black box that must be queried at finite cost. These problems are especially challenging due to: (i) the high computational cost of each evaluation (e.g., DFT, FEM, multiphysics solvers), (ii) the absence of gradients or differentiability, and (iii) the nonlinear, multimodal, and high-dimensional nature of the design landscape. In many computational design workflows, the objective is noisy, non-differentiable, and expensive to query, rendering classical gradient-based methods inapplicable. Formally, a general black-box design optimization problem can be stated as the search for a parameter $x^*$ within a bounded design domain $D^n$ that minimizes a scalar objective $f$:

\begin{equation}
    x^* = \arg\min_{x \in D^n} f(x)
\end{equation}

\noindent where the bounded domain is defined as:
\begin{equation}
    D^n = \left\{ x \in \mathbb{R}^n \;\middle|\; \mathbf{l} \leq x \leq \mathbf{u} \right\}
\end{equation}

\noindent 
Here, $n$ denotes the dimensionality of the problem, and $[\mathbf{l}, \mathbf{u}]$ defines the lower and upper bounds for each variable in the search space.  Traditionally, a broad class of algorithms has been developed to tackle high-dimensional black-box optimization, which is the core computational bottleneck in many design workflows~\cite{terayama2021black,jones1998efficient}. Early methods such as simulated annealing~\cite{pannetier1990prediction,aarts1987simulated}, minima hopping~\cite{goedecker2004minima}, metadynamics~\cite{martovnak2005simulation}, and random search~\cite{pickard2011ab} introduced probabilistic global exploration strategies that were later adopted in computational design tasks such as structure prediction, defect search, and topology refinement. Simulated annealing mimics thermal cooling to escape local minima, while minima hopping and metadynamics combine global jumps with local refinement(often combined with gradient-based local optimization), an attractive feature for physics-based design spaces with local basins.  Evolutionary algorithms (EAs), including Genetic Algorithms (GA)\cite{whitley1994genetic,oganov2010evolutionary,lambora2019genetic} and Differential Evolution (DE)\cite{chen2015measuring}, advanced this paradigm of global optimization by incorporating biologically inspired mechanisms such as selection, crossover, and mutation to evolve populations of candidate solutions over generations, enabling effective search across rugged landscapes.  In contrast to these biologically rooted methods, swarm intelligence algorithms like Particle Swarm Optimization (PSO)\cite{chen2015measuring,kennedy1997discrete,wang2010crystal} model the social behavior of decentralized agents, where each particle updates its position based on both individual experience and the collective knowledge of the swarm. This offers a simple yet powerful mechanism for balancing exploration and exploitation in large parameter spaces. Building further on nature-inspired metaheuristics, algorithms such as the Whale Optimization Algorithm (WOA)\cite{mirjalili2016whale}, Gravitational Search Algorithm (GSA)\cite{rashedi2009gsa}, and Ant Colony Optimization (ACO)\cite{dorigo2018introduction} emulate ecological and physical phenomena, ranging from prey encircling to gravitational attraction to traverse complex landscapes with adaptive behavior. Swarm intelligence methods, introduced coordination between multiple agents to balance exploration and exploitation---a principle used heavily in multi-parameter materials and engineering design problems. In parallel, Bayesian Optimization (BO)\cite{frazier2018tutorial,yamashita2018crystal}has gained traction for very expensive black-box functions by constructing probabilistic surrogate models (e.g., Gaussian processes) that are iteratively updated to guide sampling via acquisition functions, which balance uncertainty-driven exploration with exploitation of predicted optima. BO has become attractive for computational design scenarios with extremely costly evaluations (e.g., quantum calculations, CFD, FEM), by replacing direct exploration with a surrogate model and an acquisition rule to minimize the number of expensive calls.

Despite their successes, these classes of methods struggle in realistic computational design settings as dimensionality and physical constraints increase. The ``curse of dimensionality'' makes sampling sparse and uninformative, especially in multimodal landscapes with wide plateaus and nearly degenerate states---conditions ubiquitous in design spaces governed by physics. Many classical methods are also memory-less: they do not retain information from previous evaluations, leading to wasted iterations and premature convergence. These limitations have motivated the emergence of learning-augmented strategies---reinforcement learning\cite{kaelbling1996reinforcement}, surrogate-guided search\cite{chen2022radial}, and hierarchical planners\cite{chan2019machine}---that incorporate feedback from past evaluations into future sampling. For computational design, such methods provide two critical advantages: they (i) identify regions of the space that are physically promising or constraint-satisfying, and (ii) learn how to traverse the landscape adaptively rather than heuristically. Taken together, three requirements emerge for a robust computational-design optimizer:

\begin{enumerate}
\renewcommand\labelenumi{(\roman{enumi})}
    \item Adaptive sampling that scales with dimension,
    \item Mechanisms to escape shallow and non-informative regions, and
    \item Learning to exploit promising soluations in the design landscape
\end{enumerate}

In general, these allow for scalable, physically consistent, and evaluation-efficient optimization across complex scientific and engineering design problems.

\begin{figure}[h]
    \centering 
    \includegraphics[width=0.85\textwidth]{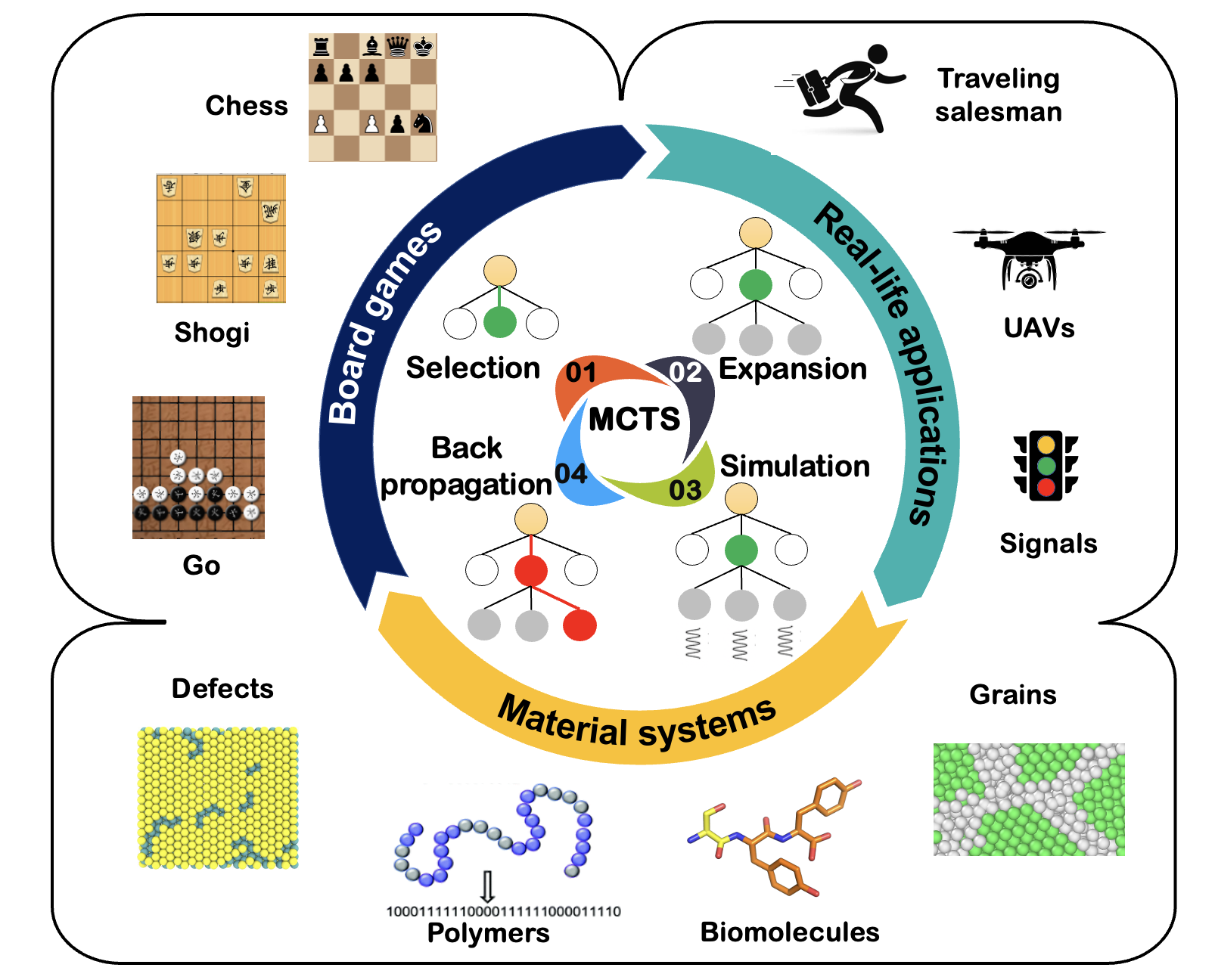}  
    \caption{
        \textit{Illustration of the breadth of Monte Carlo Tree Search (MCTS) applications, ranging from strategic decision-making in games to real-world optimization and scientific design tasks. Most existing uses operate in discrete action spaces.}
    }
    \label{fig:Figure1}
\end{figure}

Among learning-augmented strategies that aim to meet the requirements outlined above, Monte Carlo Tree Search (MCTS) provides a compelling alternative for computational design tasks that require reasoning in large search spaces without gradients. Originally developed for strategic decision-making in games such as Chess, Shogi, and Go~\cite{silver2016mastering,silver2018general}, MCTS offers a systematic way to balance exploration and exploitation without relying on handcrafted heuristics or explicit models of the objective. MCTS incrementally constructs an asymmetric decision tree using a tree policy (e.g., UCT) that prioritizes regions of the design space that are both promising and underexplored. Rollouts evaluate candidate designs to obtain a terminal score, which is then backpropagated to update the sampling policy. Unlike brute-force sampling or purely stochastic heuristics, MCTS allocates compute to informative regions of the design space, making it attractive for computational design problems where each evaluation may require a costly simulation (e.g., DFT, MD, FEM, or multiphysics solvers). Beyond board games, MCTS has been successfully applied to design and planning problems such as traveling-salesman routing, UAV navigation, autonomous perception, and combinatorial molecular and polymer design (Figure. \ref{fig:Figure1})\cite{browne2012survey,xing2020graph,rasmussen2008tree, patra2020accelerating,wang2020towards}. In materials and chemical discovery, MCTS has been used for defect structure inference\cite{banik2021learning,loeffler2021reinforcement}, grain boundary construction\cite{guziewski2020application}, exploration of combinatorial design spaces, synthesis planning\cite{wang2020towards}, drug discovery\cite{srinivasan2021artificial}, autonomous robotic\cite{kober2013reinforcement}, and hyperparameter design of simulation surrogates\cite{rakotoarison2019automated}. However, these applications are predominantly in \textit{discrete} action spaces. When MCTS is naively extended to \textit{continuous and high-dimensional computational design problems} — such as Crystal Structure Design(CSD), Interatomic Potential Model(IAP) fitting, inverse design, or multi-parameter thermo/mechanical optimization — it encounters critical limitations. Classical MCTS lacks:

(i) Adaptive sampling suited to smooth or noisy continuous objectives,
(ii) Scalability when design dimensionality grows, and
(iii) Mechanisms to learn and exploit structure in the design landscape over time.

Fixed sampling schemes that work well in discrete domains become inefficient in continuous settings, where the action space is typically unbounded and physically meaningful relations between parameters are coupled. Moreover, standard MCTS does not exploit statistical correlations or search landscape geometry accumulated during search — a crucial capability for navigating physics-constrained design spaces where useful feedback is scarce and expensive. Therefore, to extend the applicability of MCTS to continuous and high-dimensional optimization problems, substantial modifications are required. Traditional MCTS initiates exploration from a root node and expands directionally via branching mechanisms, but this structure often struggles to scale in high-dimensional settings, thus limiting its efficiency and generalization as a global optimization tool.

Our work presents a generalized framework for transforming policy-driven decision tree based reinforcement learning algorithms into efficient and scalable global optimizers. In this direction, to overcome these challenges, we propose several key innovations. First, we introduce an efficient sampling scheme combined with a direction-based learning mechanism to enable the algorithm to better propagate search trajectories across continuous, high-dimensional spaces. This improves both coverage and convergence toward optimal solutions. Second, while MCTS inherently balances exploration and exploitation through its tree policy\cite{kocsis2006bandit}, it lacks fine-grained control over the exploitation of locally promising regions, especially in complex, multimodal landscapes. Our approach enhances this capability by incorporating localized refinement strategies and adaptive reward shaping while maintaining tree consistency. Additionally, we address the local entrapment tendency of tree-based methods, where deep trees may over-explore specific subdomains, delaying global convergence. To mitigate this, we introduce a population-based extension of decision trees, incorporating the notion of global and local tree populations to simultaneously explore diverse regions of the search space while preserving depth and focus in high-value areas. We validate our framework on a suite of challenging, high-dimensional benchmark functions and demonstrate its applicability to real-world scientific and engineering problems. These include Crystal Structure Design(CSD) in materials science, the development of Interatomic Potential Model(IAP)/force fields, and continuum-scale design tasks such as pressure vessel optimization and welded joint design optimization. Across all scenarios tested, our method consistently outperforms the state-of-the-art metaheuristic and black-box optimization algorithms, demonstrating superior scalability, robustness, and convergence efficiency. Overall, this work highlights the potential of learning augmented frameworks as general-purpose optimizers for tackling complex, high-dimensional global optimization problems.

\vspace{30pt}
\begin{figure}[tp]
    \centering
    \includegraphics[width=\textwidth]{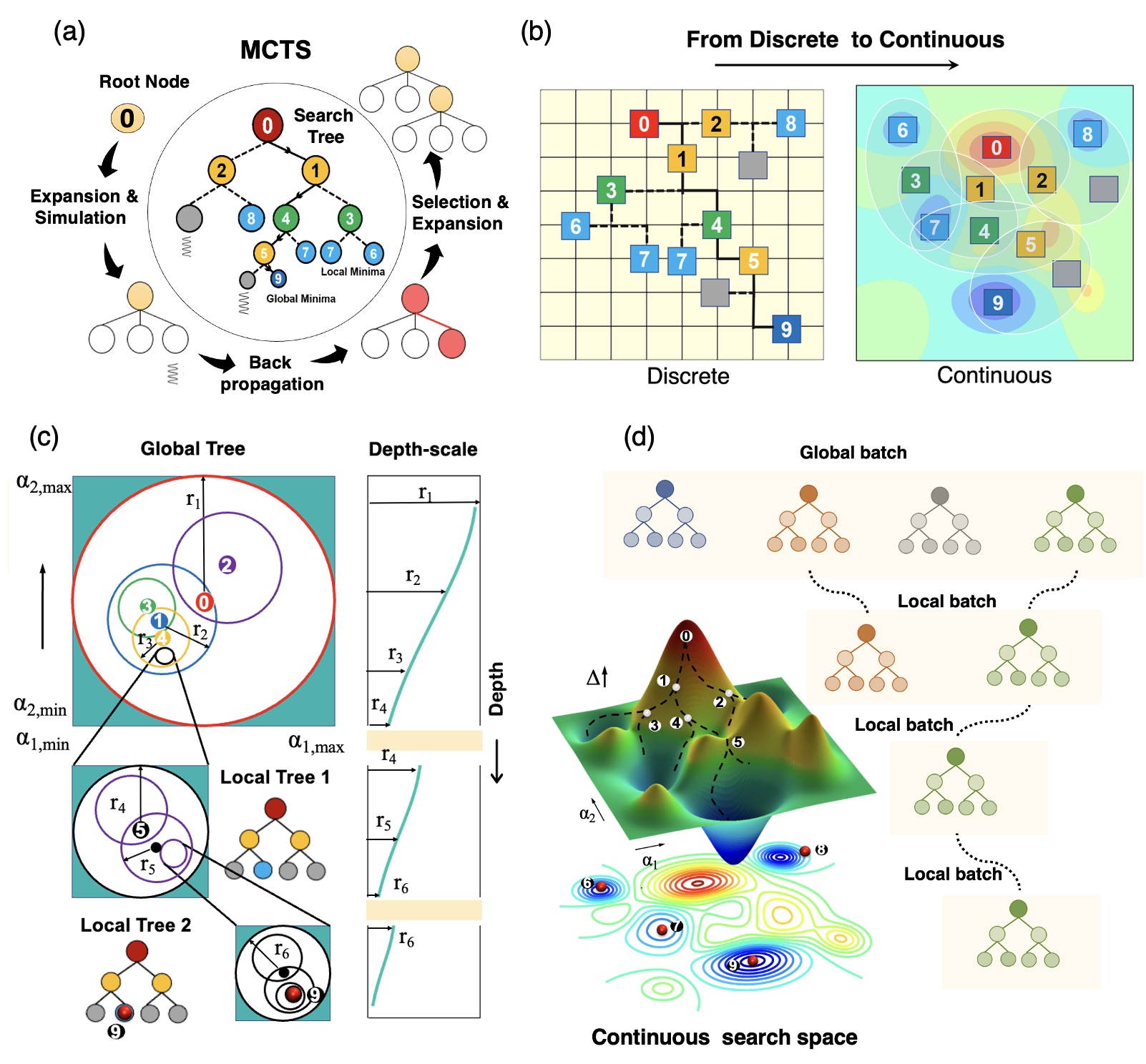}
    \caption{
        \textit{Computational design using decision tree-based MCTS for continuous search spaces. 
        (a) Different stages of MCTS, starting from a root node as the search tree grows. 
        (b) Transition from a discrete to a continuous action space, where infinitely many possible directions surround a node. 
        (c) Scaling window schemes within a tree allow deeper branches to focus exploitation near promising regions. 
        (d) A typical continuous search space illustrated by a 2D objective function. The tree typically starts from the highest objective region of the landscape and branches out through local minima. A population (batch) of trees is spawned across the continuous search space. The first stage involves global exploration; in subsequent stages, fewer trees are spawned from promising solutions for local optimization. The number of trees is progressively reduced as the search converges toward the global optimum. }
    }
    \label{fig:Figure2}
\end{figure}

\section{Methodology}

\subsection{Monte Carlo Tree Search}

The traditional Monte Carlo Tree Search (MCTS) algorithm operates through four fundamental stages: \textit{Selection}, \textit{Expansion}, \textit{Simulation}, and \textit{Backpropagation}, which are executed cyclically to explore high-dimensional, often discrete design spaces in a computationally efficient and adaptive manner (Figure~\ref{fig:Figure2}.a). This iterative structure enables MCTS to navigate vast combinatorial landscapes without relying on exhaustive enumeration.  In the \textit{Selection} phase, the algorithm traverses the current tree from the root to a leaf node by choosing child nodes that maximize an objective policy function based on the Upper Confidence Bound (UCB). This criterion balances exploitation, selecting nodes with high average rewards and exploration, visiting less-explored nodes to reduce uncertainty. The UCB for a given node $i$ is defined as follows~\cite{kocsis2006bandit}:

\begin{equation}
\text{UCB}(\text{node}_i) = \max(z_1, z_2, z_3, \dots, z_k) + C \sqrt{\frac{\ln(v_p)}{v_i}}
\label{eq:ucb}
\end{equation}

Here, $z_i$ denotes the reward obtained from previous simulations at node $i$, $v_i$ and $v_p$ represent the visit counts of the current node and its parent node respectively, and $C$ is a tunable constant that governs the exploration--exploitation trade-off.  During the \textit{Expansion} stage, a new child node is created by perturbing the configuration represented by the selected parent node. These perturbations represent discrete or quantized actions in the design space.  The \textit{Simulation} phase evaluates the newly added node through one or more rollouts (or ensemble of performance assessment of parameters in close proximity of the new node) starting from the new configuration. These rollouts can be executed using fast empirical models or higher-fidelity computational methods, with the objective of estimating relevant performance metrics associated with the candidate solution.  Finally, in the \textit{Backpropagation} stage, the rewards obtained from simulation are propagated upward through the tree, incrementally updating the statistics (e.g., visit counts and average rewards) of all ancestor nodes.

\subsection{Adaptive Sampling of Nodes in High-Dimensional Continuous Spaces}

The rollouts carried out at each node are used to sample a child node from a parent node, forming an integral part of growing and navigating the tree in the search space. In traditional MCTS, these new samples are generated using predefined discrete actions. However, in continuous-space settings (Figure. \ref{fig:Figure2}b), two major changes are necessary. First, the actions are no longer discrete but continuous in nature, requiring a mutation protocol. Second, the sampled parameters should be sufficiently spread around the node to provide a meaningful sense of directionality for selecting the next node. A key challenge arises as the dimensionality increases: evenly sampling these rollout points becomes increasingly difficult due to the exponential growth in the volume of the search space, which causes the samples to become localized. Various high-dimensional sampling schemes have been proposed to address this issue. Among the most popular are Latin Hypercube Sampling and hypersphere sampling. We utilize Latin Hypercube Sampling to generate the initial set of root nodes in the tree, ensuring a diverse set of starting points. To sample around an existing node, hypersphere sampling is employed. This involves sampling new candidate points within a $d$-dimensional hypersphere of radius $r_{\text{max}}$ centered at the current point $\mathbf{x}_r$, using:

\begin{equation}
\mathbf{x}'_r = \mathbf{x}_r + r \cdot \frac{\mathbf{u}}{\|\mathbf{u}\|}, \quad \text{where} \quad r = r_{\text{max}} \cdot \xi^{1/d}, \quad \xi \sim \mathcal{U}(0,1)
\label{eq:hypersphere_sampling}
\end{equation}

Here, $\mathbf{u} \sim \mathcal{N}(0, I)$ is a randomly drawn direction vector, and the scaling factor $r = r_{\text{max}} \cdot \xi^{1/d}$ ensures uniform sampling within the volume of the hypersphere. The random variable $\xi$ controls the radial position of the sample: drawing $\xi$ uniformly in $(0,1)$ and mapping it via $\xi^{1/d}$ yields the correct distribution of radii so that points are uniformly distributed over the hypersphere's volume (rather than concentrating near the center). One of the main limitations of hypersphere sampling is that, while it achieves isotropic spread around the node center, the number of required samples to cover the volume across a node, increases rapidly with dimensionality, leading to higher evaluations and computational cost. To overcome this limitation, we employ a direction-based sampling scheme that learns promising directions from a node and its ancestors. This model-guided approach predicts directions that are more likely to yield improved solutions, thereby significantly improving sampling efficiency in high-dimensional spaces.

\subsection{Logistic Surrogate: A Directional Sampler}

During typical sampling in tree-based optimization algorithms, each rollout at a given node is generated independently of the outcomes of previous rollouts. This memory-less nature can yield good convergence in some cases, but it has a significant limitation: if a promising direction is found during a rollout, the algorithm is unlikely to sample in that same direction again unless it explicitly generates a new parent node. However, repeated sampling in a successful direction often leads to further improvement and faster convergence. To address this, it can be beneficial to bias future samples toward directions that have previously yielded better outcomes. To address the limitations of memory-less sampling in continuous high-dimensional spaces, we implement a \textbf{Logistic Surrogate Sampler} that biases future rollouts based on the local search history of a node. For a given node parameter $x$, we collect a history of trial points and their corresponding objective outcomes, $\mathcal{H}_{\text{node}}=\{(x^{(i)}, E^{(i)})\}$. A binary classification label $y^{(i)}$ is assigned to each trial, where $y^{(i)}=1$ if $E^{(i)} < E(x)$ (success), and $0$ otherwise. The primary aspects to consider are: (i) the direction along which to sample the new point, and (ii) how far the sampled point should be from its predecessor (step size). To decouple the search for orientation from the search for step size, the relative displacement $\Delta x^{(i)} = x^{(i)} - x$ is decomposed into two distinct feature sets:
\begin{equation}
    u^{(i)} = \text{sign}(\Delta x^{(i)}) \in \{-1, 0, 1\}^d, 
    \quad \text{and} \quad 
    \phi_k\!\left(r^{(i)}\right) = \exp\!\left(-\left(r^{(i)} - c_k\right)^2\right).
\end{equation}
Here, $u^{(i)}$ represents the directional component, and $\phi(r)$ represents a feature vector where the scalar Euclidean distance $r^{(i)} = \|\Delta x^{(i)}\|$ is projected onto a basis of Gaussian Radial Basis Functions (RBFs) centered at $c_k$ across the search radius $R_{\max}$.

The algorithm employs two coupled logistic regression models to guide sampling. \textbf{Directional optimization} is performed by modeling the probability of success as
\begin{equation}
    P(y=1 \mid u) = \sigma\!\left(w_{\text{dir}}^{T}u + b\right).
\end{equation}
The trained model is then used to actively optimize the search direction $u$ using a stochastic hill-climbing procedure. Starting from the signs of the learned coefficients, the algorithm iteratively flips random dimensions of $u$, accepting changes that increase the predicted probability of success. Once the direction is fixed, \textbf{step-size sampling} is determined by the second model,
\begin{equation}
    P(y=1 \mid r) = \sigma\!\left(w_{\text{dist}}^{T}\phi(r) + b\right).
\end{equation}
To sample strictly from this learned distribution, we employ Inverse Transform Sampling. The Cumulative Distribution Function (CDF) is constructed by numerically integrating the unnormalized probability curve:
\begin{equation}
    CDF(r) = \int_{0}^{r} P(y=1\mid \rho)\, d\rho.
\end{equation}
A target value $\tau$ is drawn uniformly from $[0,\, CDF(R_{\max})]$, and the corresponding step size $r$ is recovered by solving $CDF(r) = \tau$. To bootstrap these models, the first $k$ evaluations of any node follow a fixed heuristic schedule, sampling the parent's momentum, the geometric center of the bounds, and the diagonal extrema. Subsequently, the models are retrained every $p$ iterations to adapt to the local landscape topology.

\subsection{Window Scaling for Balancing Exploration and Exploitation}

One of the crucial aspects of decision trees in continuous search spaces is managing the step size as the tree depth increases. Unlike discrete action trees, where fixed action magnitudes make every move at any depth equally spaced, in continuous spaces, this behavior can be problematic. Specifically, deeper nodes should ideally refine the search locally around promising regions, whereas shallow nodes should allow broader exploration. In other words, as the search tree grows deeper, the algorithm should gradually reduce the exploration radius to focus more precisely near optima. To address this, we implement a \textit{depth-based scaling scheme} that adaptively shrinks the sampling window size (denoted by $r_{\text{max}}$) as a function of node depth (Figure. \ref{fig:Figure2}c). The scaling factor $s$ for normalized parameters $\mathbf{x} \in [0, 1]^d$, applied to the sampling radius, is defined as:

\begin{equation}
\mathbf{s}(\text{depth}) = \mathbf{b} \cdot \exp(-\mathbf{a} \cdot \text{depth}^2)
\label{eq:depth_scaling}
\end{equation}

where:
\begin{itemize}
    \item $b \in (0, 0.5]$ is the initial scaling magnitude, typically set to 0.5,
    \item $a > 0$ controls the decay rate with depth,
    \item $\text{depth}$ is the current depth of the node in the tree.
\end{itemize}

This scaling ensures that nodes at higher depths sample smaller regions, effectively transitioning from global to local search. However, it is important to note that the appropriate scaling behavior can be task-dependent. To make the approach more robust, we typically employ \textit{multiple trees in parallel}, each initialized with a different scaling parameter $a$, commonly chosen from the range $[0.05, 0.1]$. $b$ indicate the maximum radius of the window that the search can access. These trees are grouped into \textit{global} and \textit{local} batches, with a switching protocol determining when a tree transitions from broad exploration to fine exploitation. Details of this switching mechanism and batching strategy are discussed in the following section.

\subsection{Hierarchical Batching of Tree} 

A crucial challenge with MCTS in continuous and high-dimensional spaces is that the search begins from a single point and branches out from a root node. While a single tree can effectively explore the local vicinity of its root and often converges to a nearby solution, this strategy struggles in complex, non-convex landscapes. The search tends to become overly localized, potentially getting trapped in sub-optimal regions. To escape such traps, the tree would need to grow a large number of branches to reach distant, optimal basins. This leads to exorbitant computational cost and slow convergence. To address this locality limitation, we adopt a strategy inspired by population-based methods such as evolutionary algorithms and metaheuristics. We introduce a batch of parallel trees, referred to as the \textit{global tree batch} (Figure~\ref{fig:Figure2}d). Each tree in this batch is initialized at a different point in the search space using Latin Hypercube Sampling (LHS)~\cite{minasny2006conditioned}, and is assigned a scaling parameter $a$ randomly sampled within a specified range. This ensemble enables diversified global exploration. A key feature of these global trees is that the exploration constant $C$ alternates between a fixed base value and a very large value, depending on whether the tree is making progress toward improved solutions (refer to Supplementary Figure~1). Additionally, the global trees are constrained by a maximum depth, but their width (i.e., branching factor) is unrestricted. However, their growth is conditional: a tree is allowed to expand only if it consistently yields better objective values. Once the global search phase concludes, we extract a finite subset of top-performing candidates from the trees, based on their best objective scores. These selected ones seed the next phase: the \textit{local batch}. Each local tree inherits its parent scaling factor $a_{\text{ref},i}$ and the final window size $b_{\text{ref},i}$ (from Equation~\ref{eq:depth_scaling}, where $b_{\text{ref},i} = s_{\text{final},i}$) from its corresponding global ancestor. Unlike the global phase, the local window size for the $i$-th tree in the current local batch, $b_{\text{ref},i}$, is dynamically updated during subsequent stages of newer local batch runs according to the observed improvement in the objective function in the previous stage:

\begin{equation}
b_{\text{new},i} = b_{\text{prev},i}
\left(
\frac{f_{\text{prev},i} - f_{\text{curr},i} + \epsilon}
     {f_{\text{prev},i} - f_{\text{target}} + \epsilon}
\right)^{\alpha}
\quad \text{if } f_{\text{curr},i} < f_{\text{prev},i}
\label{eq:window_expand}
\end{equation}

\begin{equation}
b_{\text{new},i} = b_{\text{prev},i} \cdot \delta
\quad \text{otherwise}
\label{eq:window_shrink}
\end{equation}

where:
\begin{itemize}
    \item $f_{\text{curr},i}$: current best objective value for tree $i$,
    \item $f_{\text{prev},i}$: previous best objective value for the same tree,
    \item $f_{\text{target}}$: target objective,
    \item $\alpha$: exponent controlling sensitivity to improvement,
    \item $\epsilon$: small constant to prevent division by zero,
    \item $\delta \in (0,1)$: decay factor applied when no improvement occurs.
\end{itemize}

This adaptive window  ensures that the accessible window of each local tree narrows as the local batch progresses (Figure~\ref{fig:Figure2}d), enabling fine-grained exploitation. Unlike the global phase, the depth of the local trees is not constrained, allowing them to grow deeper as long as they continue to improve the objective. To emphasize exploitation, we set the exploration constant to a negligible value, effectively turning each local tree into a gradient-free local optimizer focused on refining promising regions (refer to Supplementary Figure~1d--k). The local batch optimization stage is repeated for multiple rounds. After each round, non-performing trees are conditionally pruned, and only those that consistently yield improved solutions are retained. In each stage, for a given local tree, the best candidate from the previous stage is used as the new root node, ensuring continuity and retention of high-quality solutions. Importantly, after pruning, at least one best-performing tree is always retained in the batch and inherited into the next round, maintaining a persistent exploitation path throughout the search.

\vspace{30pt}
\section{Results}

\begin{figure}[tp]
    \centering
    \includegraphics[width=\textwidth]{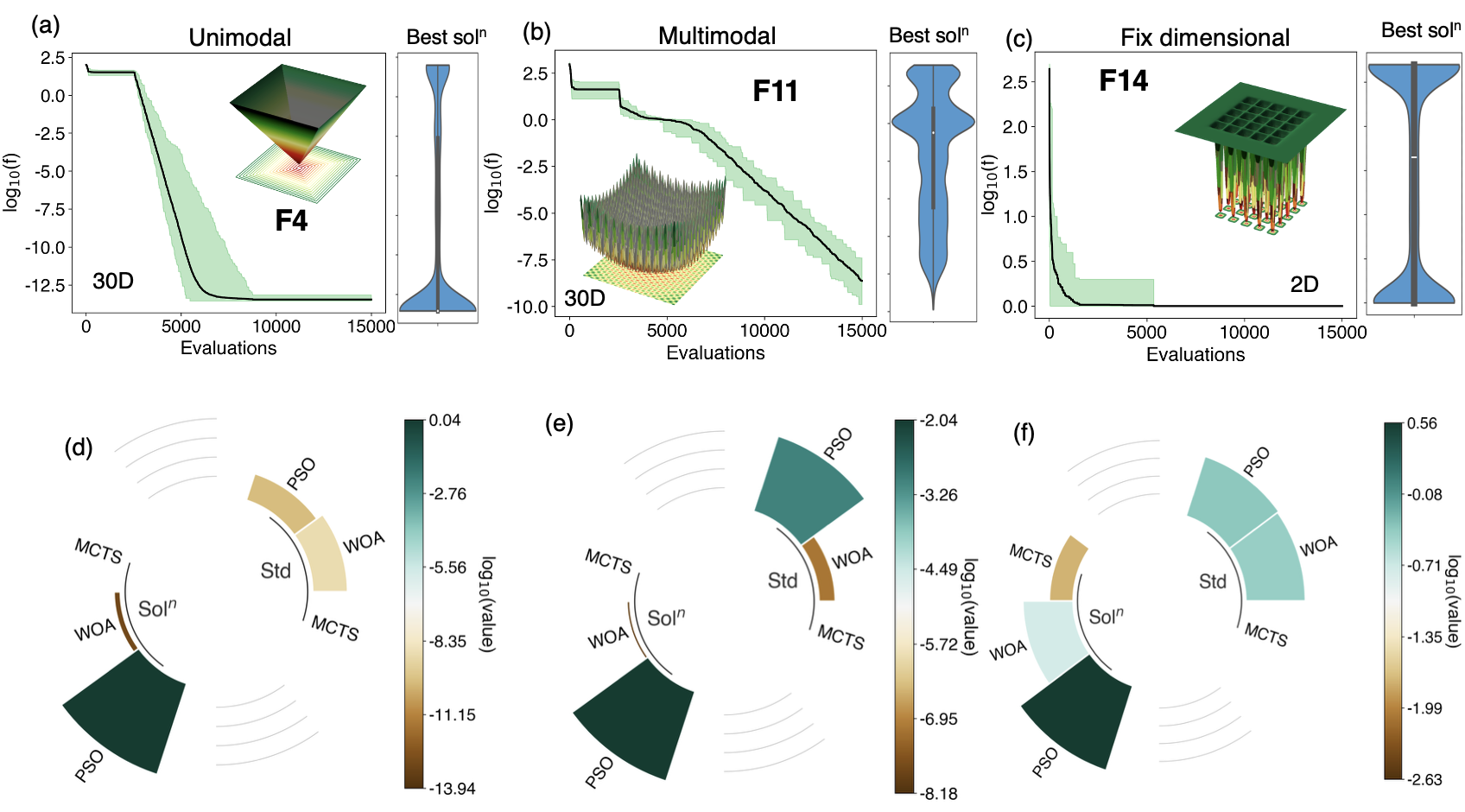}
    \caption{
        \textit{Performance Benchmarking of Our MCTS-Based Optimisation on Test Functions. (a–c) Convergence curves (logarithm of objective function  f with nuber of evaluation) for 30 independent runs, along with the sampling distribution from the search that obtained the best solution. The three benchmark functions are: Unimodal (F4), Multimodal (F11), and Fixed Dimensional (F14), respectively. (d–f) Comparison of the performance of c-MCTS, WOA, and PSO in terms of the best solutions obtained and their standard deviations over 30 independent trials for the functions shown in (a–c).}
    }
    \label{fig:Figure3}
\end{figure}

\begin{table}[h!]
\centering
\caption{Comparison of optimization results obtained for benchmark functions grouped into unimodal (F1--F7), multimodal (F8--F13), and fixed-dimensional (F14--F23) categories. The lowest average (ave) value for each function is shown in bold.}
\label{tab:trial_functions}
\renewcommand{\arraystretch}{1.4}
\setlength{\tabcolsep}{3pt}
\resizebox{\textwidth}{!}{%
\begin{tabular}{|c|cc|cc|cc|cc|cc|}
\hline
\textbf{F} & \multicolumn{4}{c|}{\textbf{MCTS}} & \multicolumn{2}{c|}{\textbf{WOA}} & \multicolumn{2}{c|}{\textbf{PSO}} & \multicolumn{2}{c|}{\textbf{Random}} \\
\cline{2-5}
 & \multicolumn{2}{c|}{\makecell{\textbf{Logistic}}} & \multicolumn{2}{c|}{\makecell{\textbf{Hypersphere}}} & \multicolumn{2}{c|}{\textbf{}} & \multicolumn{2}{c|}{\textbf{}} & \multicolumn{2}{c|}{\textbf{}} \\
\cline{2-11}
 & \textbf{ave} & \textbf{std} & \textbf{ave} & \textbf{std} & \textbf{ave} & \textbf{std} & \textbf{ave} & \textbf{std} & \textbf{ave} & \textbf{std} \\
\hline
\multicolumn{11}{|c|}{\textbf{Unimodal (UM)}} \\
\hline
F1 & \textbf{0.00000E+00} & 0.00000E+00 
   & 1.71678E--22 & 7.07787E--22 & 1.41E--30 & 4.91E--30 & 0.000136 & 0.000202 & 4.26E+04 & 3.08E+03 \\
F2 & 3.35265E--05 & 2.04925E--05 
   & 1.73077E+00 & 8.99247E+00 & \textbf{1.06E--21} & 2.39E--21 & 0.042144 & 0.045421 & 1.28E+06 & 2.37E+06 \\
F3 & \textbf{1.69045E--08} & 1.44483E--08 
   & 1.91803E+00 & 1.36430E+00 & 5.39E--07 & 2.93E--06 & 70.12562 & 22.11924 & 4.91E+04 & 5.85E+03 \\
F4 & \textbf{3.64745E--14} & 1.14179E--14 
   & 3.03847E+00 & 2.48030E+00 & 0.072581 & 0.39747 & 1.086481 & 0.317039 & 7.17E+01 & 2.60E+00 \\
F5 & \textbf{2.73319E+01} & 1.06457E--01 
   & 1.19554E+02 & 2.96610E+02 & 27.86558 & 0.763626 & 96.71832 & 60.11559 & 1.03E+08 & 1.59E+07 \\
F6 & \textbf{5.89632E--07} & 1.00714E--06 
   & 3.68666E--01 & 1.01298E+00 & 3.116266 & 0.532429 & 0.000102 & 8.28E--05 & 4.16E+04 & 3.70E+03 \\
F7 & \textbf{5.43754E--04} & 4.35580E--04 
   & 9.51802E--02 & 3.43504E--02 & 0.001425 & 0.001149 & 0.122854 & 0.044957 & 4.60E+01 & 7.16E+00 \\
\hline
\multicolumn{11}{|c|}{\textbf{Multimodal (MM)}} \\
\hline
F8 & \textbf{--7.98161E+03} & 8.17351E+02 
   & --7.62086E+03 & 5.08047E+02 & --5080.76 & 695.7968 & --4841.29 & 1152.814 & --4.16E+03 & 2.46E+02 \\
F9 & 1.42618E+00 & 7.68023E+00 
   & 1.36873E+02 & 4.22360E+01 & \textbf{0} & 0 & 46.70423 & 11.62938 & 3.42E+02 & 1.27E+01 \\
F10 & \textbf{1.42109E--15} & 2.69007E--15 
    & 1.27711E+00 & 6.91224E--01 & 7.4043 & 9.897572 & 0.276015 & 0.50901 & 1.98E+01 & 1.64E--01 \\
F11 & \textbf{6.63646E--09} & 9.64278E--09 
    & 8.04556E--03 & 8.61277E--03 & 0.000289 & 0.001586 & 0.009215 & 0.007724 & 3.84E+02 & 3.12E+01 \\
F12 & \textbf{1.04244E--05} & 9.37367E--06 
    & 1.20933E+00 & 1.62317E+00 & 0.339676 & 0.214864 & 0.006917 & 0.026301 & 1.67E+08 & 4.17E+07 \\
F13 & 6.34632E--03 & 8.94680E--03 
    & \textbf{4.09345E--03} & 5.30989E--03 & 1.889015 & 0.266088 & 0.006675 & 0.008907 & 4.06E+08 & 8.20E+07 \\
\hline
\multicolumn{11}{|c|}{\textbf{Fixed-dimensional (FD) / Composite}} \\
\hline
F14 & \textbf{9.98004E-01} & 6.00428E-11 
    & 1.031138E+00 & 1.784333E-01 & 2.111973 & 2.498594 & 3.627168 & 2.560828 & 1.0972 & 0.2789 \\
F15 & \textbf{4.17900E--04} & 2.80031E--04 
    & 1.00653E--03 & 3.03070E--04 & 0.000572 & 0.000324 & 0.000577 & 0.000222 & 0.00242 & 0.00100 \\
F16 & \textbf{--1.03163E+00} & 2.220446E-16 
    & \textbf{--1.03163E+00} & 2.22045E-16 & \textbf{--1.03163} & 4.2E--07 & \textbf{--1.03163} & 6.25E--16 & --1.02502 & 0.00598 \\
F17 & \textbf{3.97887E--01} & 0.000000E+00 
    & \textbf{3.97887E--01} & 0.000000E+00 & 0.397914 & 2.7E--05 & \textbf{0.397887} & 0 & 0.40191 & 0.00444 \\
F18 & \textbf{3.00000E+00} & 1.52550E-15 
    & \textbf{3.00000E+00} & 1.20533E-15 & \textbf{3} & 4.22E--15 & \textbf{3} & 1.33E--15 & 3.09997 & 0.08394 \\
F19 & \textbf{--3.86278E+00} & 1.42302E--13 
    & \textbf{--3.86278E+00} & 1.33227E--15 & --3.85616 & 0.002706 & \textbf{--3.86278} & 2.58E--15 & --3.85545 & 0.00531 \\
F20 & \textbf{--3.32237E+00} & 9.93810E--12 
    & \textbf{--3.32237E+00} & 9.24445E--16 & --2.98105 & 0.376653 & --3.26634 & 0.060516 & --3.06146 & 0.08211 \\
F21 & \textbf{--1.01532E+01} & 4.12262E--08 
    & --8.96974E+00 & 2.14523E+00 & --7.04918 & 3.629551 & --6.8651 & 3.019644 & --3.54978 & 0.98577 \\
F22 & \textbf{--1.04029E+01} & 4.66669E--07 
    & \textbf{--1.04029E+01} & 4.72028E--11 & --8.18178 & 3.829202 & --8.45653 & 3.087094 & --3.53778 & 0.99651 \\
F23 & \textbf{--1.05364E+01} & 1.14046E--08 
    & \textbf{--1.05364E+01} & 2.76303E--10 & --9.34238 & 2.414737 & --9.95291 & 1.782786 & --3.17910 & 0.83388 \\
\hline
\end{tabular}%
}
\end{table}

\subsection{Performance Benchmark on High Dimensional Test Functions}

Benchmark (Test) functions\cite{digalakis2001benchmarking,tang2007benchmark,molga2005test} are artificial mathematical landscapes extensively used to evaluate optimization algorithms. They come in different landscape types such as unimodal, multimodal, and fixed-dimensional multimodal. These functions are designed to stress test optimization algorithms on properties such as convergence speed, precision, robustness, scalability, and susceptibility to local minima and noise. Unimodal functions (like Sphere, Rosenbrock, Step, Schwefel 1.2) have only a single global optimum and test how rapidly and reliably an algorithm converges. Multimodal functions (like Rastrigin, Ackley, Griewank) contain many regularly or irregularly spaced local minima, challenging exploration vs. exploitation trade-offs, with difficulty increasing with dimensionality. Fixed-dimensional benchmarks (e.g., Shekel’s Foxholes, Six-Hump Camelback, Goldstein--Price, Hartmann 3D/6D, Shekel 5/7/10) operate in low dimensions (2--6), but feature dense, ``treacherous'' landscapes with deceptive features that test fine grained search behavior. Benchmark functions provide a standardized evaluation framework which allows (i) comparison of different methods under the same conditions, (ii) revelation of algorithm strengths and limitations across landscape types, (iii) assurance of reproducibility and generalization, and (iv) profiling of algorithm behavior (e.g., convergence curves, error variance across independent runs). The performance of our MCTS decision-tree-based algorithm was evaluated using a total of 23 benchmark functions (Table~\ref{tab:trial_functions}) spanning unimodal (UM), multimodal (MM), and fixed dimensional (FD) types. F$_1$--F$_7$ form the unimodal, scalable category (e.g., Sphere, Rosenbrock) with a single global optimum in a 30-dimensional search space—ideal for evaluating convergence speed and exploitation. F$_8$--F$_{13}$ are high dimensional (30-dimensional) multimodal with many local minima (e.g., Rastrigin, Ackley, Griewank), challenging global exploration and robustness. F$_{14}$--F$_{23}$ are fixed-dimensional multimodal functions (typically 2--6 dimensions). The mathematical formulation of each function is provided in Supplementary Table 1. Figure \ref{fig:Figure3} presents a representative performance analysis of our proposed algorithm alongside standard metaheuristics—Whale Optimization Algorithm (WOA) and Particle Swarm Optimization (PSO)—based on 30 independent trials for three representative functions: F$_4$ (unimodal), F$_{11}$ (multimodal), and F$_{14}$ (fixed-dimensional). For F$_4$ and F$_{11}$, our algorithm achieves numerical accuracy on the order of \(10^{-14}\) and \(10^{-9}\) respectively, while for F$_{14}$, it consistently reaches the exact global optimum (\(\sim 0.998\)), all with negligible variability near convergence. Also, looking at the sampling distribution for the individual run where the best solution is obtained (Figure~\ref{fig:Figure3}a-c), it is clear that for the unimodal function, the search initially spends some of the sampling to narrow down, and a major portion of the sampling iterations is spent on achieving high numerical accuracy—reflected in a densely populated distribution in the bottom region.

In contrast, for the multimodal function, although the initial sampling is comparable to the unimodal case, the exploration remains more uniformly distributed throughout and decreases only near convergence. For fixed dimensional functions, however the response is binary in nature. algorithm initially spends some iteration as it explores for the possible solutions, and once it finds the likely promising regions in the space, it diverts towards those promising region, as defected in the function morphology, and spends iterations in fine tuning. As shown in Figure~\ref{fig:Figure3}d--f, our algorithm not only achieves better solutions but also exhibits lower variance compared to WOA and PSO. Additional comparison across all functions is provided in Table~\ref{tab:trial_functions}. For F$_1$, the Logistic MCTS achieves the absolute zero ($0.0$), surpassing numerical precision limits and outperforming the Hypersphere baseline ($10^{-22}$). For F$_2$, WOA outperforms in numerical accuracy. For F$_3$, MCTS outperforms WOA ($1.69 \times 10^{-8}$ vs $5.39 \times 10^{-7}$), demonstrating superior fine-tuning capabilities. For F$_5$ and F$_7$, MCTS secures the best performance. From F$_8$--F$_{23}$, MCTS performs best in almost all cases. A critical result is observed on the Schwefel function (F$_8$), where the robust MCTS search achieves a mean of $-7981$, significantly deeper than the $-7620$ found by the baseline Hypersphere approach and other metaheuristics, highlighting the algorithm's ability to escape deep deceptive basins. The exception remains F$_9$ (Rastrigin), where WOA achieves a perfect $0$, while MCTS reaches $1.42$; this is probably due to the regular grid-like structure of Rastrigin which favors the specific search mechanics of WOA. Notably, for fixed-dimensional functions (F$_{14}$--F$_{23}$), MCTS consistently reaches the exact solution in all trials with numerical precision. This performance differential also provides a crucial qualitative comparison between our two sampling kernels: isotropic Hypersphere sampling versus the directional Logistic surrogate. While isotropic sampling acts as a robust fallback for noisy landscapes, the directional Logistic surrogate effectively captures local topology even in the absence of analytical gradients. As evidenced by the perfect convergence on F$_1$ and the superior basin hopping on F$_8$, the Logistic model—when implemented with robust retraining—eliminates the trade-off between exploration cost and directional accuracy, matching or outperforming the isotropic approach on 21 of the 23 benchmarks. Performance on these benchmark trends also reflect the algorithmic components introduced in Section 2. The directional logistic surrogate used at each node (Section 2.3) reduces the number of ineffective rollouts by biasing sampling toward directions that have historically improved the objective. This  can be reflected in the concentration of samples near the global optimum (Supplementary Figure~1.d-k). The depth-dependent window scaling (Section 2.4) explains why the algorithm can maintain broad global exploration in the early stages and then switch to high-precision exploitation, as evident for F$_4$ and F$_{14}$ (Figure~\ref{fig:Figure3}a-c). Finally, the global–local population of trees (Section 2.5) largely accounts for the low variance across independent runs: different trees probe different basins, and the pruning/switching protocol preferentially retains those that discover high-value regions. Together, these components help the framework remain competitive on smooth unimodal landscapes and yield a clearer advantage on rugged multimodal and fixed-dimensional benchmarks.

While benchmark test functions are designed to mimic real-world optimization problems, they often fail to fully capture the complexity of objective landscapes encountered in practical scenarios. In domains such as nanoscale materials modeling—e.g., crystal structure prediction or the development of atomistic force field models—the objective functions are typically non-convex, non-smooth energy landscapes that vary significantly with dimensionality and are further constrained by underlying physics of condensed matter. These real-world problems involve additional complexities such as symmetry constraints, chemical validity, and thermodynamic feasibility, making the optimization task far more challenging than what is reflected in idealized benchmark functions. In the following sections, we demonstrate the application of our optimization algorithm to such constrained, physical problems. We adopt standardized metrics to evaluate both the accuracy in locating global optima and the ability to respect the physical constraints inherent to the problem domain.

\begin{figure}[tp]
    \centering
    \includegraphics[width=\textwidth]{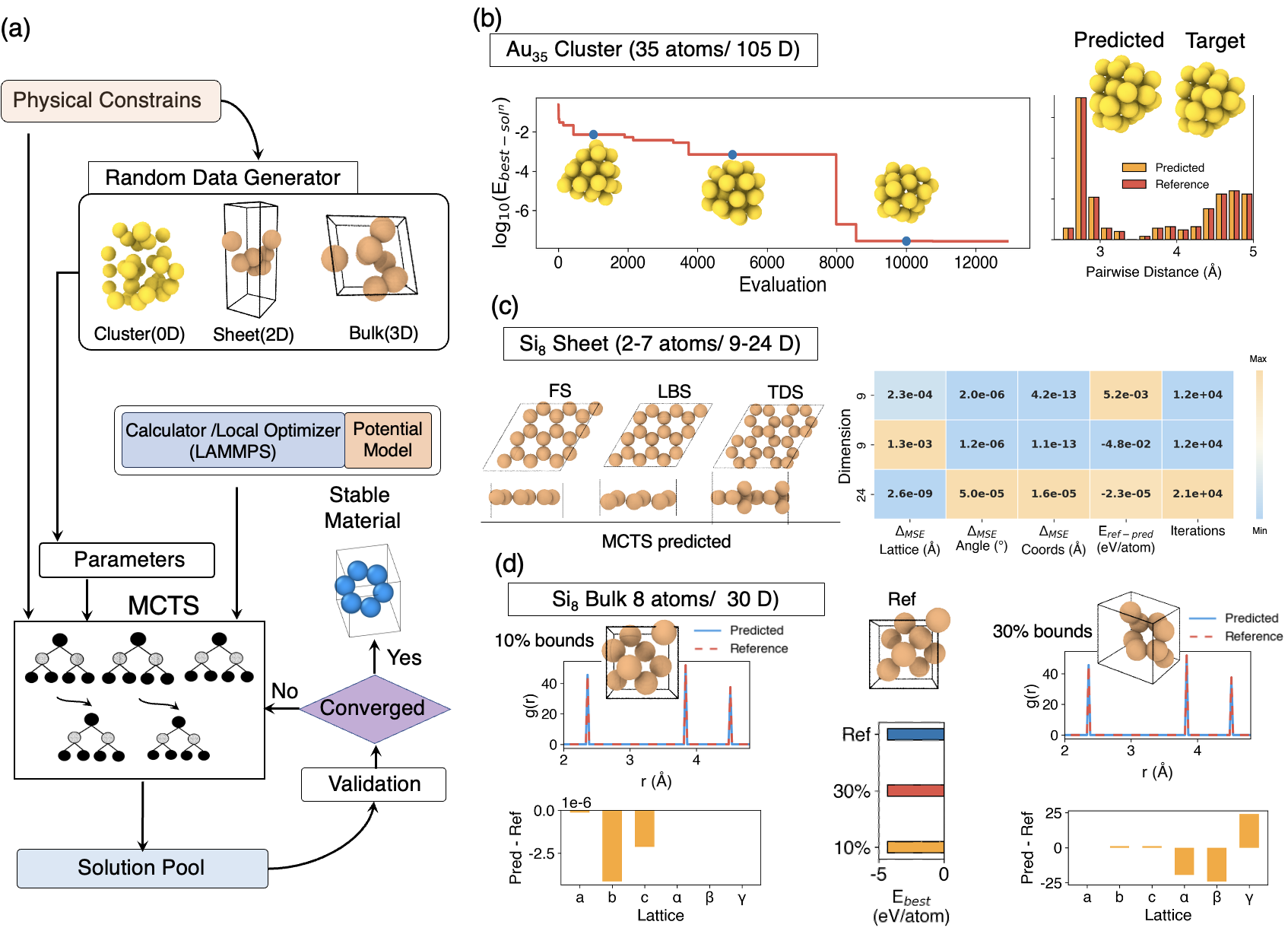}
    \caption{
        \textit{Application of our MCTS optimization algorithm in crystal structure optimization at multiple scales.
(a) The primary workflow for utilizing the algorithm in a crystal structure prediction framework, comprising a physical constraint handling module, a generator that creates physically valid configurations and converts between structural and parametric representations, the optimization algorithm, and a property evaluator or calculator.
(b) Application in predicting the global minimum of a gold (Au) nanocluster (0D material) consisting of 35 atoms (105-dimensional space), explored in terms of energy stability, and comparison of the final configuration with the reference in terms of the pairwise interatomic distance histogram.
(c) Application in sheet-like 2D materials: structure-oriented search for a metastable polymorph of silicon (silicene), targeting configurations with specific structural attributes. The unit cell predicted by MCTS, as well as the prediction accuracy across different polymorphs, is shown.
(d) Assessment of performance in optimizing the periodic bulk (3D material) ground-state structure of silicon (cubic diamond), starting from an amorphous, high-energy configuration using an energy-based objective search. The final obtained configuration, the reference configuration, and their structural comparison, in terms of the RDF, are shown for different bounds (ranges of the search space) (10\% and 30\%).}
    }
    \label{fig:Figure4}
\end{figure}

\subsection{Multiscale Crystal Structure Design}

The physical properties of materials are intrinsically linked to the arrangement of atoms in their motifs and the interplay between local order and long-range interactions. At the atomic level, these interactions manifest as a complex energy landscape with countless local minima, representing stable and metastable sahses of materials. Many of these states exhibit unique properties—chemical, thermal, optical, mechanical, and electronic—that have shown promising potential. To obtain materials belonging to a specific phase within a specific composition, the structural variability must be accounted for. Structural variability shapes the energy landscape, influencing stability and symmetry, thereby guiding the search for new materials. By creating diverse energy basins, structural variability allows for the existence of both stable and metastable configurations\cite{balasubramanian2025machine,balasubramanian2024learning}. From an algorithmic perspective, algorithms that explore the energy landscape perceive it as a set of quantitative parameters to be tuned and explored\cite{banik2023continuous,oganov2019structure}. These explorations must be guided with proper chemical constraints. However, typical material landscapes grow exponentially with number of atoms. For example, a system with $N$ atoms has a minimum of $3N$ dimensions accounting for its Cartesian coordinates, while additional parameters may come from composition and the lattice vectors. Higher dimensions lead to exponential growth in the search space as more atoms or degrees of freedom are added, resulting in challenges in identifying low-energy configurations. A major bottleneck is the computational requirement associated with evaluating the stability of this material or the properties using some calculator or evaluator model. Higher fidelity models offer more accuracy, but come at a higher computational cost. Thus, it necessitates the exploration algorithm to be evaluation-efficient and capable of reaching the target with the minimum possible iterations. Another challenge in optimizing this physical system is the degeneracy. For example, to navigate the search space one must introduce some changes to the parameters or the coordinates or the lattice parameters of the crystals\cite{banik2023continuous,atahan2014prediction,oganov2011modern}. However, due to symmetric constraints, often small changes lead to configurations that are representatives of the same energy basin. This causes over-sampling in local energy basins, wasting a lot of precious iterations of the optimizer. Thus, it needs the mutations (i.e., the changes introduced) to be sufficient enough to make the search exploratory, but not too much to make it random. This balance between exploration and controlled mutation is built in through our window-scaling scheme and the directional logistic surrogate introduced in Section 2.3. For this application, we use hypersphere sampling as depicted in Equation~\ref{eq:hypersphere_sampling}, treating $r=1$ to sample on the surface of the hypersphere and then using the learned directional weights to bias the proposals towards promising regions. Depending on the application, either the lattice parameters or the coordinates are perturbed individually as a randomly selectable action to sample new nodes in the search tree.

In Figure. \ref{fig:Figure4}.a, the basic workflow for utilizing our search algorithm for optimizing the material system at nanoscale is illustrated. The basic workflow has four different components. One component which is crucial is the physical constraint block, which handles all the necessary composition and lattice parameter, interatomic distance constraints, and physical symmetry of the system. This block also defines the search space, the ranges of lattice parameters, the number of atoms, and the composition parameters, defining the bounds. The second block is the random data generator, which creates random atomistic configurations within the given bounds obeying the physical constraints. This also converts configuration into mutable parameters. The third block is the MCTS, where the optimization or sampling of new candidates happens based on the methods discussed earlier. And the fourth block, which is the calculator(objective calculation) block or evaluator block. Typically atomistic optimization\cite{thompson2022lammps} with Interatomic potential models(IAPs)\cite{banik2024development,manna2022learning}, quantum calculations using density functional theory (DFT)\cite{hafner2008ab} are used . This block primarily contributes in the computation of stability of material, whether in terms of energy or the property which is used as an objective to optimize. These blocks also serve as a local optimizer, which is quite useful during the sampling as one would want to sample stable candidates. The local optimizer helps the distorted samples to push to local energy basins and helps obtain a relatively stable version of the candidate. Post-search, we analyze all the candidates and their stability and structural traits to determine the convergence of the workflow. The corresponding search space constraints used for the problems explored are described in Supplementary Table 2.

We demonstrate three diverse problems in the domain of structure optimization using our algorithm. First, we consider the search for the global minimum of gold (Au) nanoclusters. Nanoclusters, as zero-dimensional material systems, have applications across various domains\cite{chakraborty2017atomically}. Au nanoclusters are of significant interest due to their versatile applicability\cite{arvizo2010gold}. Computationally, the global minima of such clusters across different sizes have been extensively explored\cite{doye1998global}, making this system an ideal benchmark for structure optimisation problems. In this work, we employ the Sutton--Chen\cite{todd1993surface} interatomic potential to recover the known global minimum of a 35-atom Au nanocluster. For nanocluster systems, the only dimensionality arises from atomic coordinates, resulting in a 105-dimensional optimization problem, which is sufficiently large.  As shown in Figure. \ref{fig:Figure4}b, the search converges in under 12,000 evaluations—a decent result given the high dimensionality ($>$100D).  For comparison, Doye and Wales~\cite{doye1998global} performed global optimization of Sutton--Chen metal clusters up to $N = 80$ atoms using a basin-hopping (Monte Carlo minimization) scheme. For each cluster size, they employed five independent runs of 5000 Monte Carlo steps, i.e., $25{,}000$ local minimizations per system, with each step involving a complete local geometry optimization on the transformed landscape. The figure also illustrates the evolution of atomic configurations throughout the search. We used the LAMMPS\cite{thompson2022lammps} molecular dynamics package to locally optimize sampled configurations at each node, guiding them toward the nearest local energy basin. The Sutton--Chen potential was employed with the LAMMPS calculator. The final obtained configuration shows an exact match with the known global minimum in terms of pairwise interatomic distances, confirming the accuracy of our search in recovering the ground-state structure.

Silicon’s crucial role in modern electronics has earned it the title of the ``Silicon Age.'' Among its emerging forms, silicene\cite{grazianetti2016two}—a two-dimensional (2D) allotrope structurally analogous to graphene—has gained significant interest due to its potential in next-generation technologies, including flexible electronics and quantum computing. As a polymorphic material, silicene exhibits diverse structural phases\cite{leoni2021demonstration,mazdziarz2023transferability} that can challenge conventional structure optimisation methods. Multiple polymorphs of silicene are known, including flat silicene (FS), low-buckled silicene (LBS), trigonal dumbbell silicene (TDS), honeycomb dumbbell silicene (HDS), and large honeycomb dumbbell silicene (LHDS). These structures are not only geometrically diverse but also differ significantly in terms of energy stability, atomic arrangements, and lattice parameters. The metastable nature of these configurations makes them challenging to recover through direct energy-based searches, as they represent sub-optimal basins in the energy landscape which a typical global optimizer is likely to overlook or bypass entirely. \textbf{Flat silicene (FS):} 2 atoms, 9-dimensional space (6 Cartesian coordinates + 3 lattice parameter, a,b,$\gamma$), \textbf{Low-buckled silicene (LBS):} 2 atoms, 9-dimensional space, and \textbf{Trigonal dumbbell silicene (TDS):} 7 atoms, 24-dimensional space. The goal is to recover each target unit cell configuration starting from a randomly initialized atomic arrangement, using the following objective function, that measure the structural similarity:

\[
\text{Score} = \frac{1}{D} \left\| \mathbf{l}^{\text{pred}} - \mathbf{l}^{\text{ref}} \right\|_1 + \frac{1}{D} \left\| \mathbf{r}^{\text{pred}} - \mathbf{r}^{\text{ref}} \right\|_1,
\]

\noindent where
\[
\begin{aligned}
&\mathbf{l}^{\text{pred}},\, \mathbf{l}^{\text{ref}} \in \mathbb{R}^{n_{\text{lattice}}} \quad &&\text{(predicted and reference lattice vectors, e.g., } a, b, c, \alpha, \beta, \gamma\text{)} \\
&\mathbf{r}^{\text{pred}},\, \mathbf{r}^{\text{ref}} \in \mathbb{R}^{3N} \quad &&\text{(flattened atomic coordinates, centered.)} \\
&D = n_{\text{lattice}} + 3N \quad &&\text{(total dimensionality of the parameter vector)}, \quad n_{\text{lattice}} \in [0, 6]
\end{aligned}
\]

It should be noted that here we are not using stability (energy) as a metric, as searches based on stability will push the search toward the most stable polymorph. We are rather interested in finding the local metastable configurations. Figure.~\ref{fig:Figure4}c showcases the final predicted polymorphs by our workflow for three representative polymorphs. Since the search is driven by the score mentioned above, we independently calculate its energy using LAMMPS + MEAM\cite{du2011energy} potential model. From the trend of evolution of the best score with the number of iterations (refer to Supplementary Figure. 2a-c), for most of the cases there is moderate to high coordination between the subsequent evolution of the energy. For FS, an energy saddle point is obtained far before the score converges to the optimum value, indicating degeneracy - solutions with the same energetics may give internally different configurations. This is also true as FS is energetically unstable compared to LBS, although they share a similar unit cell and an identical search space was used. Thus, a purely energy-driven search would likely end up in LBS. This trend can also be seen in its objective evolution, since both score and energetics tend to converge simultaneously. A similar trend to FS can again be seen for TDS. Overall, as can be seen in Figure.~\ref{fig:Figure4}c, the predicted structures from our search closely resemble the target reference configurations. The lattice parameter errors are on the order of $10^{-4}$ on average, while the lattice angles and atomic coordinates are on the order of $10^{-5}$. We do observe a slight increase in the error magnitude as the dimensionality increases from 9 dimensions to 24 dimensions. The iterations taken for the 9-dimensional case were approximately $\sim 12{,}000$, and for the 24-dimensional case, approximately $\sim 21{,}000$. Although all the searches converged to the exact solution, it should be noted that in this case we did not use any local optimization. This is purely a parametric search with the objective of obtaining the desired parameters.

Next, we look at obtaining the global minimum of bulk silicon, which is the cubic diamond configuration as shown in Figure.~\ref{fig:Figure4}d. In this case, similar to the nanocluster system, we use local optimization with LAMMPS + Stillinger Weber\cite{stillinger1985computer} potential model to relax the configurations sampled at new nodes. Here, the dimensionality is 30, with 24 Cartesian coordinates and six lattice parameters( a,b,c, $\alpha$,$\beta$, $\gamma$). An essential aspect of this search for the bulk system is the bounds of the search space, specifically in terms of the lattice vector range. Allowing more bounds enables the search to obtain configurations with similar stacking and energy, but diverse in lattice parameters. We use two bounds, \(\pm10\%\) and \(\pm30\%\) around the reference parameters (Supplementary Table 2), to conduct the search. As shown in Figure.~\ref{fig:Figure4}d, for the 10\% bound, the MCTS obtains the exact solution of the unit cell both in terms of RDF and lattice vectors. For the 30\% bound, it obtains a cubic diamond stacking, but with different lattice.

\begin{figure}[tp]
    \centering
    \includegraphics[width=\textwidth]{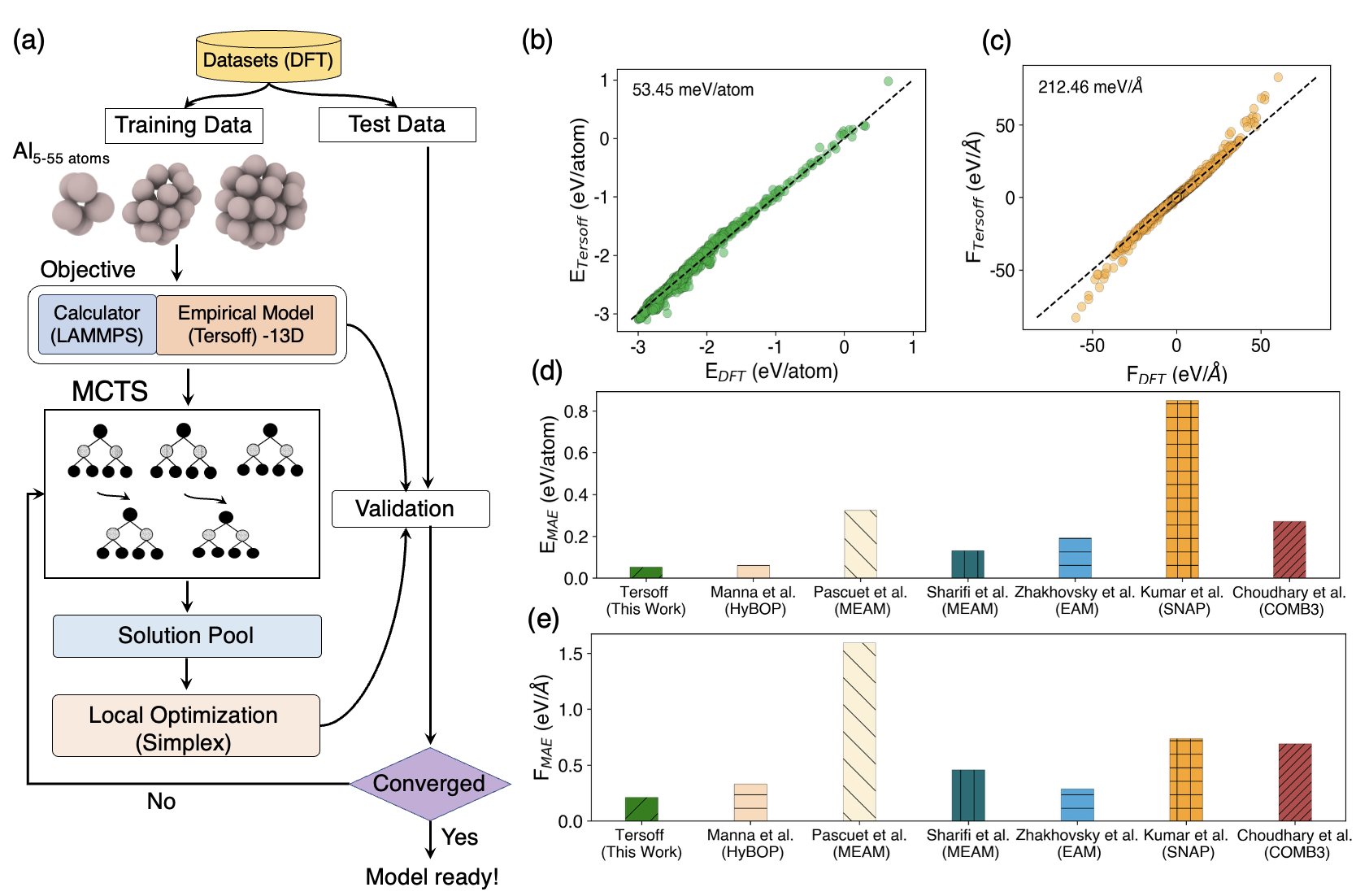}
    \caption{
        \textit{Application of the MCTS algorithm for optimizing potential energy models. (a) Workflow illustrating the application of the MCTS-based optimization framework for developing potential energy models. The process includes four major components: (i) generation of a reference dataset from DFT or other high-fidelity methods, (ii) a calculator that maps the predicted model parameters to physical material properties, (iii) the MCTS optimization engine, and (iv) a local optimizer used for fine-tuning model parameters. (b,c) Parity plots comparing the predicted energies and forces with reference DFT data for aluminum (Al) nanoclusters, along with associated prediction errors on the test dataset. (d,e) Performance benchmarking of the developed model against standard Al potentials from the literature in terms of energy and force prediction errors.}
    }
    \label{fig:Figure5}
\end{figure}

\subsection{Computational Design of High-Dimensional Potential Energy Models}

In the previous section, crystal structure search served as a prototypical forward computational design task — selecting structural candidates and evaluating their stability under physical constraints. A second, equally central design challenge arises one level deeper in the pipeline: designing the surrogate physics model itself (the Interatomic Atomic potential (IAP) models). In practice, crystal search, phase stability screening, and autonomous discovery all depend on an interatomic potential that must be designed to mimic high-fidelity energetics with minimal cost. Empirical potentials serve exactly this design role: they are compact analytic functions constructed to approximate quantum energetics and forces, enabling design loops to operate at realistic scales without invoking DFT at every iteration\cite{stillinger1985computer,du2011energy,manna2022learning,pereyaslavets2022accurate}. However, constructing such a potential is itself a high-dimensional inverse design optimization problem: one must search a rugged parameter space to find a parameterization whose predictions reproduce a target reference landscape. This becomes particularly complex in nanoclusters, where the underlying energy landscape is highly multimodal, populated by near-degenerate isomers separated by shallow barriers.

Among available model families, the Tersoff bond-order potential is a widely used design architecture due to its physics-informed structure: it embeds local coordination and angular geometry into a dynamic bond-order term, allowing it to represent diverse chemistries (C, Si, SiO\textsubscript{x}, ZrO\textsubscript{2}) with a compact analytic form\cite{tersoff1988new,manna2022learning}. Its locality and simplicity make it computationally efficient enough to be placed inside a design loop while retaining sufficient physical fidelity. Motivated by the benchmark study of Manna et al.\cite{manna2022learning}, we treat the parameterization of a Tersoff-type model for Al nanoclusters as a second use case of our computational design framework — not to design a structure, but to design the potential itself such that it reproduces DFT-derived energetics and forces from the same dataset. In doing so, we demonstrate that our method is not limited to structural design, but applies equally to the design of surrogate models that enable scalable simulation-driven discovery.

The second use case therefore evaluates the proposed method not on structural discovery itself, but on a core meta-design task — learning the potential energy model that will, in turn, govern downstream computational design workflows. The basic workflow for utilizing our search algorithm for development of a prototypical potential model based on Tersoff formalism is shown in Figure.~\ref{fig:Figure5}a. The workflow primarily comprises four main sections: the first is the dataset generated from DFT or high-fidelity techniques; the second is a calculator that utilizes the predicted model parameters to map the properties of physical materials; the third is the optimization algorithm itself; and the fourth is a local optimizer for fine-tuning the models.  The training dataset and the test data are the nanoclusters of elemental Al, with size distribution ranging from 5--55 atoms. For additional details on the dataset, refer to the work by Manna et al.\cite{manna2022learning}. We used the LAMMPS ASE interface as an evaluator to calculate the energy and forces during optimization. The objective function is basically a weighted linear combination of energy and force MSE errors used during the fitting.  From multiple searches, we select best solution pools based on performance on the test set and further fine-tune them using a simplex\cite{dantzig2016linear} algorithm. The performance of the final model is shown in Figure.~\ref{fig:Figure5}(b--c). The model achieves an energy error of 53.45~meV/atom and a corresponding force error of 212.46~meV/\AA, which is better than the state-of-the-art HYBOP (Tersoff + correction) developed specifically for nanoclusters. Additionally, we compare the performance of our model with standard models available in the literature~[cite], such as MEAM\cite{pascuet2015atomic,sharifi2025developing}, EAM\cite{zhakhovskii2009molecular}, SNAP\cite{kumar2023transferable}, and COMB\cite{choudhary2014charge}.  As shown in Figure.~\ref{fig:Figure5}(d,e), the developed model performs exceptionally well compared to the other counterparts. Most traditional models either fail to capture a tight correlation between predicted and DFT energies and forces, or exhibit large systematic errors in the high-energy regime, where their predictions deviate strongly from the DFT diagonal (Supplementary Figure 3). This indicates the efficacy of our algorithm in finding a solution that is able to capture physical properties of this material in a high-dimensional landscape. The dimension of the search was 13, and the obtained best set of parameters is provided in Supplementary Table 3.

\begin{figure}[h]
    \centering
    \includegraphics[width=\textwidth]{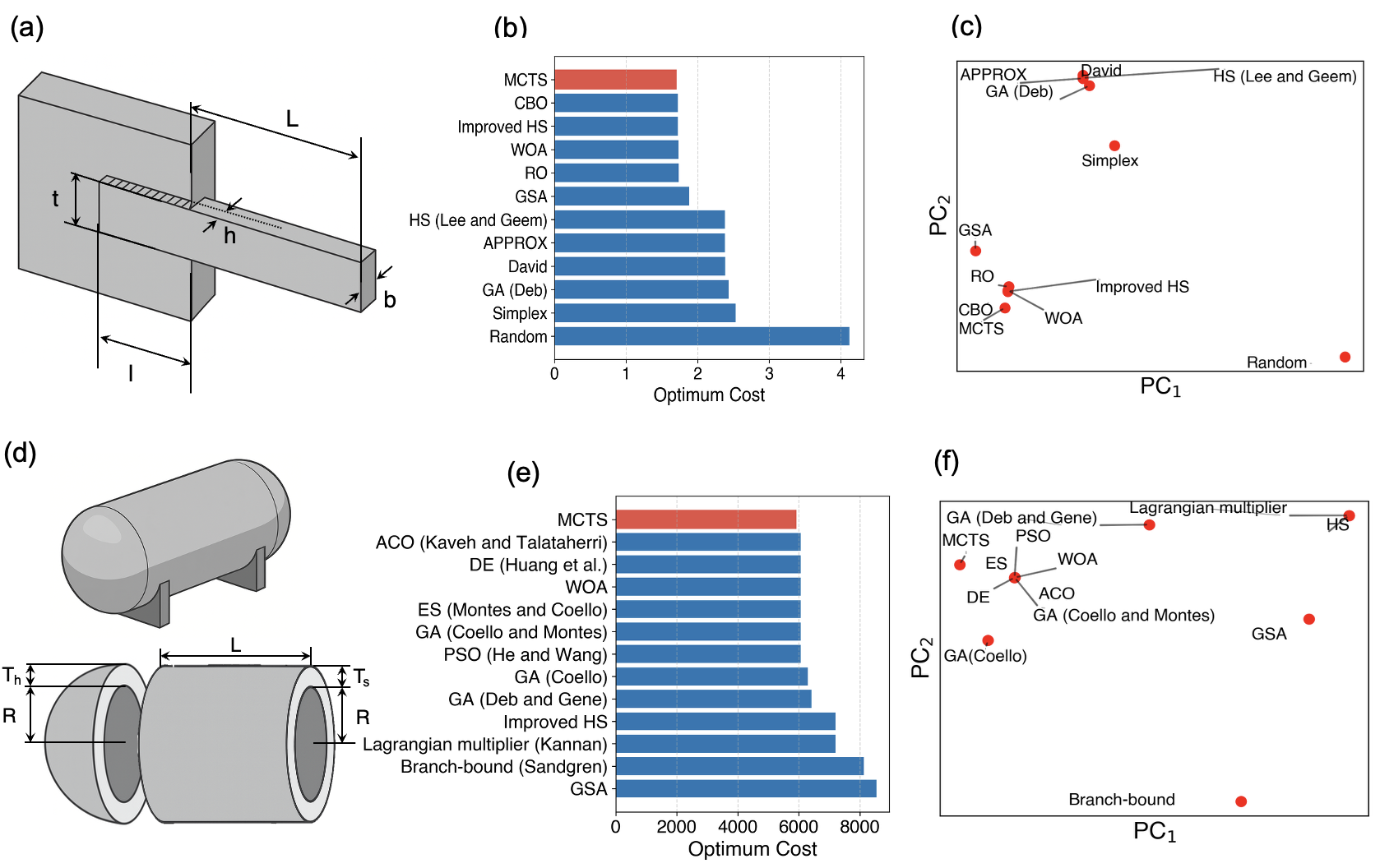}
    \caption{
        \textit{Application of the MCTS algorithm to continuum scale design optimization problems.
(a) Design optimization of a welded joint, with the schematic illustrating key geometric parameters subject to optimization. (b) Comparative performance of the MCTS algorithm against 11 optimizers reported in the literature. (c) Reduced-dimensional principal component analysis (PCA) representation of the optimal solutions obtained from different optimizers for the welded joint problem. (d) Schematic of a pressure vessel showing 11 geometric design parameters used in the optimization. (e) Performance benchmarking of the MCTS algorithm against other optimizers for the pressure vessel design problem. (f) Reduced-dimensional PCA plot of optimized parameter sets obtained from different optimizers for the pressure vessel design.}
    }
    \label{fig:Figure6}
\end{figure}

\subsection{Continuum-Scale Modeling for Design Optimization Problems}

Continuum-scale engineering design problems are archetypal constrained optimization tasks, where feasibility and performance are jointly enforced by physics and fabrication limits. To illustrate our method in this regime, we apply it to two standard benchmark design problems — a welded beam and a pressure vessel — each governed by strict mechanical and geometric constraints. As in the previous applications, samples violating constraints are penalized, steering the search away from infeasible regions of the design space. The formulation of these problems and the corresponding optimization results are presented below.

\vspace{10pt}
\noindent
\textbf{Design Optimization of Welded Beam}

\noindent

The welded-beam problem is a canonical benchmark in constrained engineering design: the objective is to minimize the total fabrication cost under mechanical constraints on shear stress $(\tau)$, bending stress $(\sigma)$, critical buckling load $(P_c)$, and tip deflection $(\delta)$. The design is parameterized by four continuous geometric variables --- weld thickness $h$, weld length $l$, bar height $t$, and bar thickness $b$ (Figure.~\ref{fig:Figure6}a) --- which directly govern both cost and structural performance. For compact notation we set $x_1=h$, $x_2=l$, $x_3=t$, and $x_4=b$. The governing constraint equations and admissible bounds are provided in Supplementary Section~1. We evaluate our method on this four-dimensional design space against standard optimizers reported in the literature\cite{mirjalili2016whale}, using 30 independent trials to assess reproducibility. As seen in Figure.~\ref{fig:Figure6}b, the algorithm consistently converges to nearly identical optima, with only modest variability in early iterations ($\sim$ first 6000 steps) before all runs collapse to a common solution after the exploitation phase; detailed convergence traces are provided in Supplementary Figure 4a. The best design identified by our method (score = 1.697958) outperforms all competing approaches (Figure.~\ref{fig:Figure6}b) . A reduced-dimensional comparison of the resulting parameter sets ((Figure.~\ref{fig:Figure6}c) shows that only the top two performers (MCTS and CBO) arrive at similar regions of the design space, whereas other methods converge to distinctly separated clusters --- revealing the presence of multiple feasible minima and underscoring the difficulty of constrained computational design problems.

\vspace{10pt}
\noindent
\textbf{Pressure Vessel Design}

The second constrained design problem considered is the cylindrical pressure vessel\cite{mirjalili2016whale}, whose objective is to minimize total fabrication cost---including material, forming, and welding---subject to geometric and physical constraints on shell thickness, head thickness, allowable volume, and maximum vessel length. The design is defined by four continuous parameters: the shell thickness $T_s$, head thickness $T_h$, inner radius $R$, and the cylindrical length without the head $L$ (Figure.~\ref{fig:Figure6}d). These parameters directly determine both manufacturability and structural feasibility. For compact notation, we set $x_1 = T_s$, $x_2 = T_h$, $x_3 = R$, and $x_4 = L$. The detailed constraint formulation is provided in \textcolor{red}{Supplementary Section~2}.

This problem is again four-dimensional, and we benchmark the proposed method against standard algorithms previously applied to this design task\cite{mirjalili2016whale}. As shown in Figure.~\ref{fig:Figure6}e, we report the best solution obtained over 13 independent runs. The solutions produced by our method are again highly consistent across trials—MCTS repeatedly attains very similar best costs across independent runs (Figure.~\ref{fig:Figure6}e), with detailed convergence traces provided in Supplementary Figure 4b. Interestingly, the reduced-dimensional projections of the resulting parameters reveal that MCTS converges to regions of the design space distinct from those explored by competing algorithms---indicating that the method is capable of uncovering feasible yet previously unexplored optima.

Overall, the welded-beam and pressure-vessel studies demonstrate that the proposed approach generalizes beyond atomistic domains and is effective for constrained continuum-scale design problems. The method not only identifies feasible and high-quality optima, but also explores alternative design regions that remain unreachable to conventional optimizers, highlighting its suitability for computational design under complex constraints.

\begin{table}[H]
\centering
\caption{Best parameters and corresponding scores obtained by MCTS for the welded beam and pressure vessel design problems.}
\begin{tabular}{|c|c|c|c|c|}
\hline
\textbf{Problem} & \textbf{Variable} & \textbf{Symbol} & \textbf{Best Value} & \textbf{Best Score} \\
\hline
\multirow{4}{*}{Welded Beam} 
    & Weld thickness & \( h \) & 0.204508 & \multirow{4}{*}{1.697958} \\
    & Welded joint length & \( l \) & 3.273933 & \\
    & Bar height & \( t \) & 9.046498 & \\
    & Bar thickness & \( b \) & 0.205730 & \\
\hline
\multirow{4}{*}{Pressure Vessel} 
    & Shell thickness & \( T_s \) & 0.779536 & \multirow{4}{*}{5898.135917} \\
    & Head thickness & \( T_h \) & 0.385230 & \\
    & Inner radius & \( R \) & 40.332212 & \\
    & Cylinder length & \( L \) & 199.959890 & \\
\hline
\end{tabular}
\label{tab:mcts_best_params}
\end{table}

\section{Discussion}

Our results across various benchmarks and three representative yet distinct design regimes demonstrate that the proposed physics-informed tree search framework can act as a general optimization layer for computational design. Unlike classical metaheuristics that rely on fixed sampling heuristics, or prediction-oriented machine-learning models that do not actively decide where to search, the method presented here performs policy-driven reallocation of computation toward statistically and physically informative regions of the landscape. This capability proved essential in atomistic structure search, in the inverse design of potential energy models, and in constrained continuum-scale engineering problems, all of which operate at different scales but share the same three characteristics: expensive evaluations, non-smooth and multimodal landscapes, and the absence of reliable gradients.

The performance gains observed on benchmark functions are not merely numerical improvements but consequential for design workflows: the method exhibits low variance across runs, preserves feasibility when constraints are present, and remains robust when dimensionality increases. In atomistic settings---where search spaces are large and degeneracy is pervasive---the ability to adaptively refine sampling windows and bias directions based on learned statistics enabled convergence with far fewer evaluations than baseline approaches. In inverse potential fitting---a problem rarely framed as a design task---the method successfully navigated a high-dimensional parameter landscape to produce a surrogate model outperforming established alternatives. In continuum design---a domain dominated by constraint-driven solution spaces---the method consistently uncovered feasible optima, including regions unexplored by competing algorithms, underscoring its capacity for global coverage under constraint filtering.

Placed in the broader context of AI for scientific discovery, our work reflects a shift from prediction to decision-making. Recent advances in deep learning and foundation models have accelerated the ability to represent materials, encode priors, and interpolate known physics, but the core bottleneck in scientific design remains the selection of what to evaluate next under strict cost and feasibility budgets. Standard RL and MCTS frameworks have historically flourished in discrete environments, but fail to extend directly to continuous, high-dimensional physics-constrained design spaces. The present framework bridges that gap by incorporating direction learning, depth-aware exploitation, and population-level branching into a single policy that respects both search geometry and domain constraints. In this sense, it acts not as a post-hoc optimizer but as a \emph{decision layer that sits inside the design loop} of computational science.

Looking forward, tree-based reinforcement paradigms of this kind may serve as the missing connection between simulation engines, surrogate models, and autonomous experimentation---allowing AI to not only learn from scientific data but to act upon scientific objectives in a targeted and data-efficient manner.

\subsection{Limitations and Outlook}

Despite its generality and strong performance, the framework has several limitations that outline concrete directions for extension. First, the approach operates in a strict black-box regime and does not yet exploit multi-fidelity structure or partial differentiability where available. Second, knowledge gained in one problem is not amortized across related tasks; trees do not share learned structure across runs, preventing transfer learning across design families. Third, feasibility is currently imposed through penalties rather than through constraint-aware generative models or symbolic feasibility priors. Promising extensions include: (i) foundation-guided search in which pretrained physical encoders constrain exploration, (ii) multi-agent and memory-sharing tree ensembles that accumulate structural priors across tasks, and (iii) closed-loop integration with simulation and experiment, where the same policy governs data acquisition and refinement.

\subsection{Perspective}

Our work illustrates how learning-guided tree search can elevate optimization from a downstream numerical routine to the central organizing mechanism of computational design. By embedding active decision-making into the loop of simulation, physical constraint handling, and surrogate evaluation, the approach exemplifies a shift toward AI-native design frameworks---systems that do not merely model scientific spaces, but navigate them strategically. As computational physics, machine learning, and autonomous experimentation converge, such decision-centric optimization layers are poised to become foundational infrastructure for scientific discovery.

\section{Acknowledgments}

This is based upon work supported by the U.S. Department of Energy, Office of Science, Office of Basic Energy Sciences Data, Artificial Intelligence, and Machine Learning at DOE Scientific User Facilities program under Award Number 34532 (Digital Twins). This work was performed in part at the Center for Nanoscale Materials, which is a U.S. Department of Energy of Science User facilities supported by the U.S. Department of Energy, Office of Science, Office of Basic Energy Sciences, under Contract No. DE-AC02-06CH11357. This work utilized the National Energy Research Scientific Computing Center, a DOE Office of Science User Facility  supported by the Office of Science of the U.S. Department of Energy under Contract No. DE-AC02-05CH11231. Dr. Banik would also like to thank the Advanced Cyberinfrastructure Coordination Ecosystem\cite{boerner2023access}: Services \& Support (ACCESS) program, which is supported by U.S. National Science Foundation grants \#2138259, \#2138286, \#2138307, \#2137603, and \#2138296.

\section{Author Contributions}
S.B. and S.K.R.S.S. conceived the project. T.D.L. and S.K.R.S.S. adapted the MCTS algorithm for continuous search. S.B. performed all calculations, validation, and analysis. S.B. and S.K.R.S.S. wrote the manuscript. T.D.L. developed the logistic surrogate. S.B. adapted the algorithm for population-based search utilizing local and global schemes. S.B. developed the crystal structure prediction framework and continuum modeling framework. S.B. developed the potential model development workflow with input from H.C. and S.M. O.Y. provided feedback on the code repository and GitHub demonstrations. All authors discussed the results and provided valuable feedback on the manuscript. S.K.R.S.S. supervised the overall project. The authors also thank Dr. Anirban Chandra for useful discussions.

\section{Data and Code Availability}

The Python implementation of the physics-informed MCTS framework, \texttt{PyPhysTree}, which incorporates global and local tree batching, adaptive hypersphere sampling, and the logistic surrogate, is available at \href{https://github.com/sbanik2/PyPhysTree}{\url{https://github.com/sbanik2/PyPhysTree}}. The repository includes workflows, datasets, and benchmark scripts for unimodal, multimodal, and composite optimization functions, as well as example demonstrations for crystal structure prediction, interatomic potential evaluation, and continuum design problems. While this repository enables full reproducibility of the methodology, the specific robust implementation of the original MCTS engine with logistic search used to generate the majority of the production results in this work is available from the authors upon reasonable request. The original crystal structure prediction specifically relied on the CASTING framework \href{https://doi.org/10.1038/s41524-023-01128-y} {\url{https://doi.org/10.1038/s41524-023-01128-y}}.

\newpage
\bibliography{references}
\bibliographystyle{unsrt}
\end{document}


\title{Supplementary Information: Physics-Informed Tree Search for High-Dimensional Computational Design} 



\author[1,2,*]{Suvo Banik}
\author[1,2]{Troy D. Loeffler}
\author[1]{Henry Chan}
\author[1,2]{Sukriti Manna}
\author[3]{Orcun Yildiz}
\author[3]{Tom Peterka}
\author[1,2,*]{Subramanian Sankaranarayanan}

\affil[1]{Department of Mechanical and Industrial Engineering, University of Illinois, Chicago, Illinois 60607, United States}
\affil[2]{Center for Nanoscale Materials, Argonne National Laboratory, Lemont, Illinois 60439, United States}
\affil[3]{Mathematics and Computer Science Division, Argonne National Laboratory, Lemont, Illinois 60439, United States}
\affil[*]{Corresponding authors: \href{mailto:skrssank@uic.edu}{skrssank@uic.edu}, \href{mailto:sbanik2@anl.gov}{sbanik2@anl.gov};}

\maketitle

\begin{table}[H]
\caption{Description of benchmark functions.}
\renewcommand{\arraystretch}{2.2}
\normalsize
\begin{adjustbox}{width=\textwidth}
\begin{tabular}{|p{11.5cm}|p{1.2cm}|p{1.8cm}|p{1.6cm}|}
\hline
\textbf{Function} & \textbf{V\_no} & \textbf{Range} & $f_{\min}$ \\
\hline
$F_1(x) = \sum_{i=1}^{n} x_i^2$ & 30 & $[-100, 100]$ & 0 \\
$F_2(x) = \sum_{i=1}^{n} |x_i| + \prod_{i=1}^{n} |x_i|$ & 30 & $[-10, 10]$ & 0 \\
$F_3(x) = \sum_{i=1}^{n} \left( \sum_{j=1}^{i} x_j \right)^2$ & 30 & $[-100, 100]$ & 0 \\
$F_4(x) = \max_i\{|x_i|\}$ & 30 & $[-100, 100]$ & 0 \\
$F_5(x) = \sum_{i=1}^{n-1} \left[100(x_{i+1} - x_i^2)^2 + (x_i - 1)^2\right]$ & 30 & $[-30, 30]$ & 0 \\
$F_6(x) = \sum_{i=1}^{n} \left(\lfloor x_i + 0.5 \rfloor \right)^2$ & 30 & $[-100, 100]$ & 0 \\
$F_7(x) = \sum_{i=1}^{n} i x_i^4 + \text{random}[0,1)$ & 30 & $[-1.28, 1.28]$ & 0 \\
$F_8(x) = \sum_{i=1}^{n} -x_i \sin(\sqrt{|x_i|})$ & 30 & $[-500, 500]$ & $-418.9829 \times 30$ \\
$F_9(x) = \sum_{i=1}^{n} \left[x_i^2 - 10\cos(2\pi x_i) + 10\right]$ & 30 & $[-5.12, 5.12]$ & 0 \\
$F_{10}(x) = -20 \exp\left(-0.2 \sqrt{\frac{1}{n} \sum x_i^2}\right) - \exp\left(\frac{1}{n} \sum \cos(2\pi x_i)\right) + 20 + e$ & 30 & $[-32, 32]$ & 0 \\
$F_{11}(x) = \frac{1}{4000} \sum x_i^2 - \prod \cos\left(\frac{x_i}{\sqrt{i}}\right) + 1$ & 30 & $[-600, 600]$ & 0 \\
\parbox[t]{10.3cm}{
$F_{12}(x) = \frac{\pi}{n} \left[10 \sin^2(\pi y_1) + \sum_{i=1}^{n-1} (y_i - 1)^2 (1 + 10 \sin^2(\pi y_{i+1})) + (y_n - 1)^2 \right] + \sum u(x_i,10,100,4)$\\
$y_i = 1 + \frac{x_i + 1}{4}, \quad u(x,a,k,m) = \begin{cases}
k(x - a)^m & x > a \\
0 & |x| \le a \\
k(-x - a)^m & x < -a
\end{cases}$
} & 30 & $[-50, 50]$ & 0 \\
\hline
\end{tabular}
\end{adjustbox}
\end{table}

\setcounter{table}{0} 

\begin{table}[H]
\caption{Description of benchmark functions (continued).}
\renewcommand{\arraystretch}{2.2}
\normalsize
\begin{adjustbox}{width=\textwidth}
\begin{tabular}{|p{11.5cm}|p{1.2cm}|p{1.8cm}|p{1.6cm}|}
\hline
\textbf{Function} & \textbf{V\_no} & \textbf{Range} & $f_{\min}$ \\
\hline
\parbox[t]{11.5cm}{
$F_{13}(x) = 0.1 \left[\sin^2(3\pi x_1) + \sum_{i=1}^{n-1} (x_i - 1)^2 (1 + \sin^2(3\pi x_{i+1})) \right.$\\
$\left. + (x_n - 1)^2 (1 + \sin^2(2\pi x_n))\right] + \sum u(x_i,5,100,4)$
} & 30 & $[-50, 50]$ & 0 \\
$F_{14}(x) = \left(\frac{1}{500} + \sum_{j=1}^{25} \sum_{i=1}^{2} \frac{1}{(x_i - a_{ij})^6} \right)^{-1}$ & 2 & $[-65, 65]$ & 1 \\
$F_{15}(x) = \sum_{i=1}^{11} \left[a_i - \frac{x_1(b_i^2 + b_ix_2)}{b_i^2 + b_ix_3 + x_4} \right]^2$ & 4 & $[-5, 5]$ & 0.00030 \\
$F_{16}(x) = 4x_1^2 - 2.1x_1^4 + \frac{1}{3}x_1^6 + x_1x_2 - 4x_2^2 + 4x_2^4$ & 2 & $[-5, 5]$ & $-1.0316$ \\
$F_{17}(x) = \left(\frac{x_1 - \frac{5.1}{4\pi^2}x_1^2 + \frac{5}{\pi}x_1 - 6}{4}\right)^2 + \left(1 - \frac{1}{8\pi}\right) \cos(x_1) + 10$ & 2 & $[-5, 5]$ & 0.398 \\
\parbox[t]{11.5cm}{
$F_{18}(x) = \left[1 + (x_1 + x_2 + 1)^2 (19 - 14x_1 + 3x_1^2 - 14x_2 + 6x_1x_2 + 3x_2^2)\right]$\\
$\times \left[30 + (2x_1 - 3x_2)^2 (18 - 32x_1 + 12x_1^2 + 48x_2 - 36x_1x_2 + 27x_2^2)\right]$
} & 2 & $[-2, 2]$ & 3 \\
$F_{19}(x) = - \sum_{i=1}^{4} c_i \exp\left(-\sum_{j=1}^{3} a_{ij}(x_j - p_{ij})^2\right)$ & 3 & $[1, 3]$ & $-3.86$ \\
$F_{20}(x) = - \sum_{i=1}^{4} c_i \exp\left(-\sum_{j=1}^{6} a_{ij}(x_j - p_{ij})^2\right)$ & 6 & $[0, 1]$ & $-3.32$ \\
$F_{21}(x) = \sum_{i=1}^{5} \left[(X - a_i)(X - a_i)^T + c_i\right]^{-1}$ & 4 & $[0, 10]$ & $-10.1532$ \\
$F_{22}(x) = \sum_{i=1}^{7} \left[(X - a_i)(X - a_i)^T + c_i\right]^{-1}$ & 4 & $[0, 10]$ & $-10.4028$ \\
$F_{23}(x) = \sum_{i=1}^{10} \left[(X - a_i)(X - a_i)^T + c_i\right]^{-1}$ & 4 & $[0, 10]$ & $-10.5363$ \\
\hline
\end{tabular}
\end{adjustbox}
\end{table}

\begin{figure}[H]
    \centering
    \includegraphics[width=\textwidth]{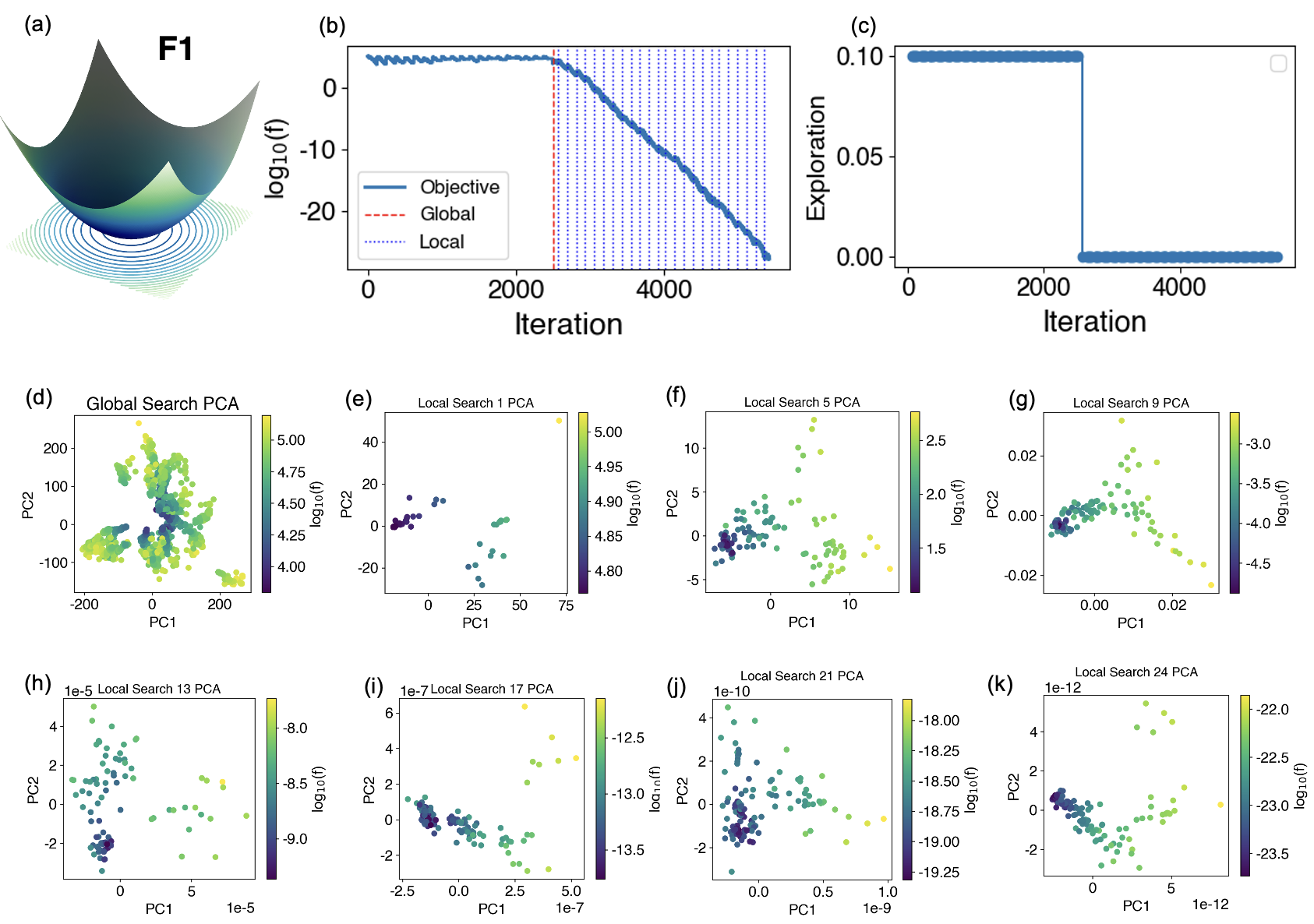}
    \caption{\textit{Exploration of the search space of the F1 Sphere function (30D) using hierarchical batching of the trees. (a) The representative Sphere function in three dimensions. (b) Sequential best objective evolution in global and local trees as the search progresses and converges toward the solution. (c) Adaptive adjustment of the exploration constant during the search; once the search enters the local regime, the exploration constant is reduced to a negligible value to promote exploitation. (d–k) Reduced-dimensional PCA plots of the sampled parameters as the search progresses through different stages of the global and local trees. Each point represents a parameter vector, color-coded by its objective value.}}
    \label{fig:FigureS1}
\end{figure}

\begin{table}[H]
\centering
\caption{Constraints for Structure Optimization Problems}
\renewcommand{\arraystretch}{1.4}
\begin{tabular}{|l|c|c|c|c|}
\hline
\textbf{System} & \textbf{Composition} & \textbf{$n_{\text{atom}}$} & \textbf{Lattice Vectors (\AA)} & \textbf{Lattice Angles ($^\circ$)} \\
\hline
Au nanocluster       & Au (1)   & [35, 35]  & [20, 20]    & [90, 90]  \\
Silicene (FS)        & Si (1)   & [2, 2]    & [2.67, 4.96] & [84, 156] \\
Silicene (LBS)       & Si (1)   & [2, 2]    & [2.67, 4.96] & [84, 156] \\
Silicene (TDS)       & Si (1)   & [7, 7]    & [4.5, 8.4]   & [84, 156] \\
Si Bulk (10\% bounds) & Si (1)  & [8, 8]    & [4.9, 5.9]   & [81, 99]  \\
Si Bulk (30\% bounds) & Si (1)  & [8, 8]    & [3.7, 7.0]   & [63, 117] \\
\hline
\end{tabular}
\label{tab:constraints}
\end{table}

\begin{figure}[H]
    \centering
    \includegraphics[width=\textwidth]{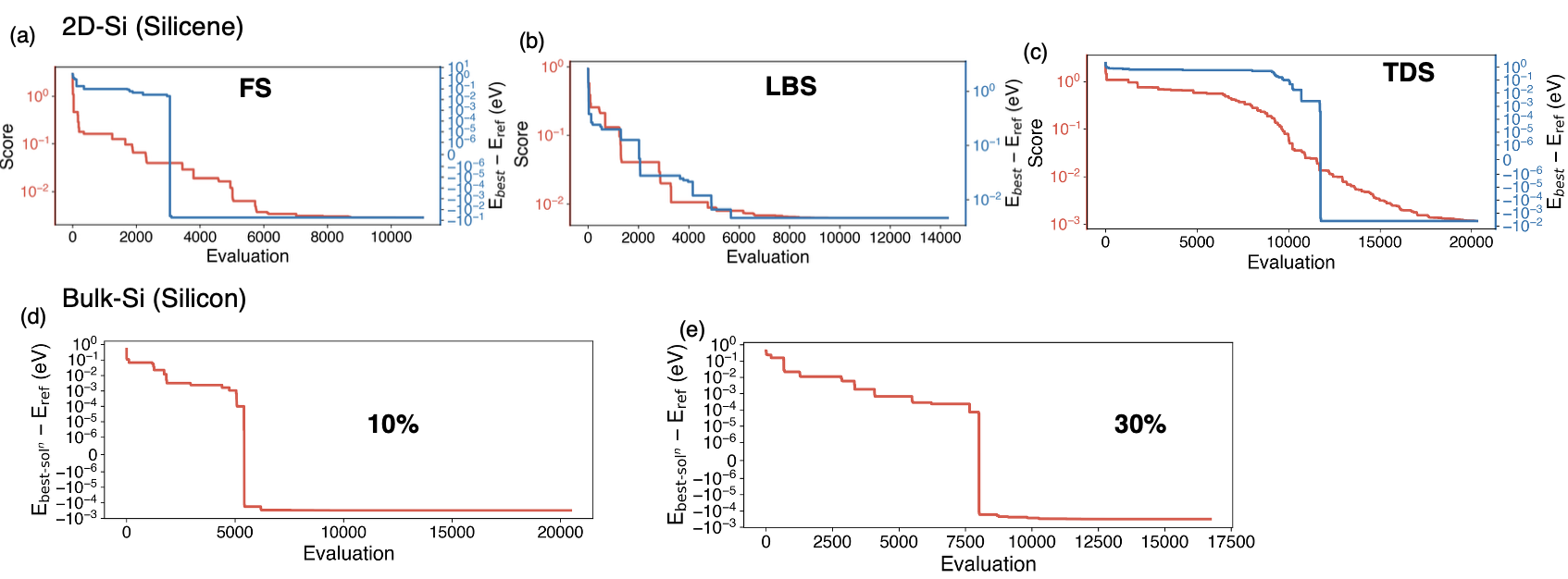}
    \caption{\textit{Evolution of the score/objective function with evaluations for crystal structure optimization problems. (a–c) Evolution of the best objective score and corresponding energy difference from the reference polymorphs (stability values calculated independently) for polymorphs of silicene: flat silicene (FS), low-buckled silicene (LBS), and trigonal dumbbell silicene (TDS), respectively. (d,e) Evolution of the objective function (cohesive energy) and its difference from the reference polymorph (cubic diamond) as the search progresses, for two different bounds: 10\% and 30\%.}}
    \label{fig:FigureS1}
\end{figure}

\begin{table}[H]
\centering
\caption{Tersoff potential parameters for Al Nanoclusters.}
\renewcommand{\arraystretch}{1.6}
\begin{tabular}{|c|c|c|}
\hline
\textbf{Parameter} & \textbf{Value} & \textbf{Status} \\
\hline
$i$ (Element 1) & Al & Fixed \\
$j$ (Element 2) & Al & Fixed \\
$k$ (Element 3) & Al & Fixed \\
$m$  & 1 & Fixed \\
$\gamma$ & 0.08558915042412107 & Optimized \\
$\lambda_3$ & 1.388924075919019 & Optimized \\
$c$ & 102609.51092476014 & Optimized \\
$d$ & 505.9045930232983 & Optimized \\
$h$ & -1.597491319465445 & Optimized \\
$n$ & 1.0675452081168926 & Optimized \\
$\beta$ & 2.6268160450036113 & Optimized \\
$\lambda_2$ & 0.946808360148141 & Optimized \\
$B$ & 32.614504273907656 & Optimized \\
$R$  & 8.701216493495025 & Optimized \\
$D$  & 0.6982431003188023 & Optimized \\
$\lambda_1$ & 2.631369725037805 & Optimized \\
$A$ & 785.7021664508125 & Optimized \\
\hline
\end{tabular}
\label{tab:tersoff_Al}
\end{table}

\begin{figure}[H]
    \centering
    \includegraphics[width=\textwidth]{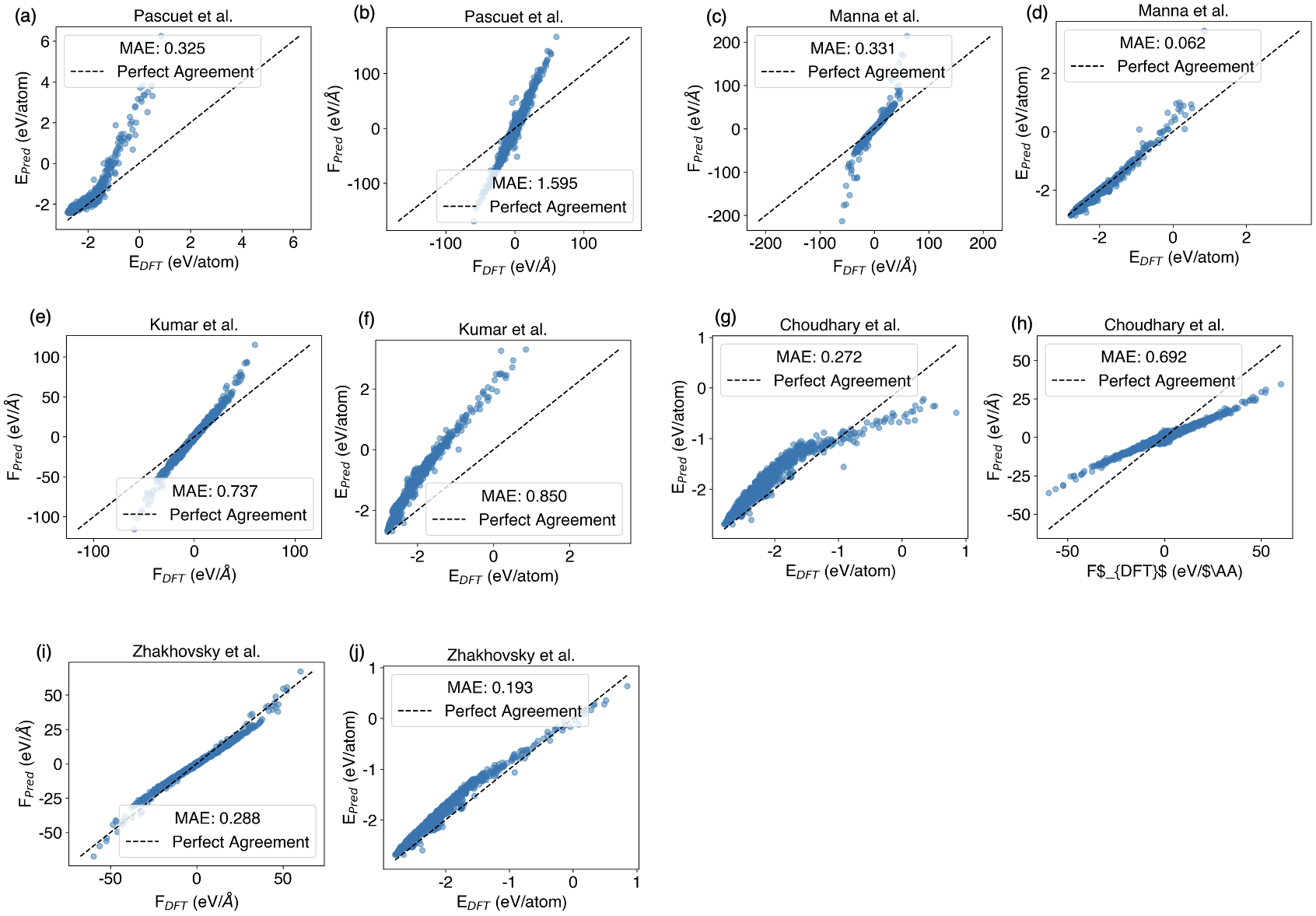}
    \caption{\textit{Performance of potential models in the literature for Al in predicting the energy and forces of nanocluster systems. (a–j) Energy and force parity plots along with the MAE values for different models used in the main manuscript for performance evaluation and comparison.}}
    \label{fig:FigureS1}
\end{figure}

\section{Design Optimization of Welded Beam}
\begin{itemize}
    \item \( x_1 = h \): weld thickness (\text{in}),
    \item \( x_2 = l \): length of the welded joint (\text{in}),
    \item \( x_3 = t \): height of the bar (\text{in}),
    \item \( x_4 = b \): thickness of the bar (\text{in}).
\end{itemize}

\vspace{1em}
\noindent\textbf{Objective Function:}
\[
\text{Minimize:} \quad f(x) = 1.10471 x_1^2 x_2 + 0.04811 x_3 x_4 (14 + x_2)
\]

\vspace{1em}
\noindent\textbf{Subject to Constraints:}

\[
\begin{aligned}
g_1(x) &= x_1 - x_4 \leq 0 & \text{(Weld thickness not greater than bar thickness)} \\
g_2(x) &= \tau(x) - 13600 \leq 0 & \text{(Shear stress limit)} \\
g_3(x) &= \sigma(x) - 30000 \leq 0 & \text{(Bending stress limit)} \\
g_4(x) &= 0.125 - x_1 \leq 0 & \text{(Minimum weld thickness)} \\
g_5(x) &= \delta(x) - 0.25 \leq 0 & \text{(End deflection limit)} \\
g_6(x) &= P - P_c(x) \leq 0 & \text{(Buckling load constraint)}
\end{aligned}
\]

\vspace{1em}
\noindent\textbf{Intermediate Expressions:}

\[
\begin{aligned}
\tau &= \sqrt{(\tau')^2 + 2\tau' \tau'' \frac{l}{2R} + (\tau'')^2} \\
\tau' &= \frac{6000}{\sqrt{2} x_1 x_2} \\
\tau'' &= \frac{6000 (14 + \frac{x_2}{2}) \sqrt{0.25(x_2^2 + (x_1 + x_3)^2)}}{2 J} \\
J &= 2 \sqrt{2} x_1 x_2 \left( \frac{x_2^2}{12} + \left( \frac{x_1 + x_3}{2} \right)^2 \right) \\
\sigma &= \frac{504000}{x_3^2 x_4} \\
\delta &= \frac{65 \cdot 6000 \cdot 14^3}{30 \cdot E x_3^4 x_4} \\
P_c &= \frac{4.013 E \sqrt{\frac{x_3^2 x_4^6}{36}}}{L^2} \left( 1 - \frac{x_3}{2L} \sqrt{\frac{E}{4G}} \right)
\end{aligned}
\]

\vspace{1em}
\noindent\textbf{Material Constants:}

\[
\begin{aligned}
E &= 30 \times 10^6 \ \text{psi} \\
G &= 12 \times 10^6 \ \text{psi} \\
P &= 6000 \ \text{lbf} \\
L &= 14 \ \text{in}
\end{aligned}
\]

\begin{figure}[H]
    \centering
    \includegraphics[width=\textwidth]{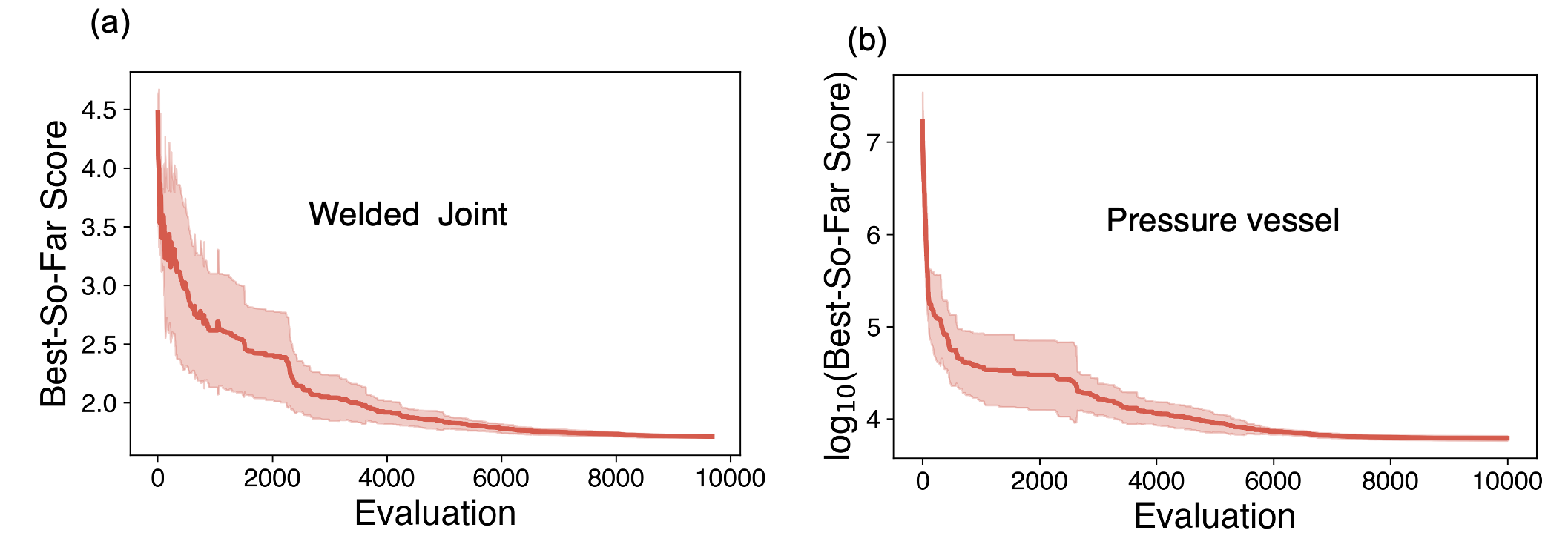}
    \caption{\textit{Convergence traces of the proposed algorithm for design optimization problems. (a) shows the performance and evolution of the objective for the welded joint optimization problem, and (b) shows the performance for the pressure vessel design problem.}}

\end{figure}

\vspace{1em}
\noindent\textbf{Design Variable Bounds:}

\[
\begin{aligned}
0.1 \leq x_1, x_4 \leq 2.0 \\
0.1 \leq x_2, x_3 \leq 10.0
\end{aligned}
\]

\vspace{1em}
\noindent The objective is to find the design vector \( x = [x_1, x_2, x_3, x_4] \) that minimizes the fabrication cost while satisfying all mechanical and geometric constraints.

\section{Design Optimization of Pressure Vessel}

\begin{itemize}
    \item \( x_1 = T_s \): thickness of the shell (in),
    \item \( x_2 = T_h \): thickness of the head (in),
    \item \( x_3 = R \): inner radius (in),
    \item \( x_4 = L \): length of the cylindrical section (in).
\end{itemize}

\noindent
\textbf{Objective Function:}
\[
\text{Minimize: } f(x) = 0.6224x_1x_3x_4 + 1.7781x_2x_3^2 + 3.1661x_1^2x_4 + 19.84x_1^2x_3
\]

\noindent
\textbf{Subject to Constraints:}
\[
\begin{aligned}
g_1(x) &= -x_1 + 0.0193x_3 \leq 0 \quad &\text{(Shell thickness constraint)} \\
g_2(x) &= -x_2 + 0.00954x_3 \leq 0 \quad &\text{(Head thickness constraint)} \\
g_3(x) &= -\pi x_3^2 x_4 - \frac{4}{3} \pi x_3^3 + 1,\!296,\!000 \leq 0 \quad &\text{(Volume constraint)} \\
g_4(x) &= x_4 - 240 \leq 0 \quad &\text{(Length constraint)}
\end{aligned}
\]

\noindent
\textbf{Design Variable Bounds:}
\[
\begin{aligned}
0 \leq x_1, x_2 \leq 99 \\
10 \leq x_3, x_4 \leq 200
\end{aligned}
\]

\noindent
The objective is to find the design vector \( x = [x_1, x_2, x_3, x_4] \) that minimizes the total cost of the pressure vessel while satisfying all geometric and volume constraints.